\documentclass{article}

\PassOptionsToPackage{numbers, compress}{natbib}

\usepackage[preprint]{neurips_2026}
\usepackage[utf8]{inputenc}   
\usepackage[T1]{fontenc}      
\usepackage{microtype}        

\usepackage{amsmath}
\usepackage{amssymb}
\usepackage{amsfonts}         
\usepackage{nicefrac}         
\usepackage{bm}

\usepackage{booktabs}         
\usepackage{multirow}         
\usepackage{makecell}         
\usepackage{longtable}        
\usepackage{tabu}             
\usepackage{colortbl}         

\usepackage{graphicx}         
\usepackage{subcaption}       
\usepackage{wrapfig}          
\usepackage[export]{adjustbox}
\usepackage{rotating}         
\usepackage{capt-of}          

\usepackage{xcolor}           
\usepackage{tcolorbox}        
\usepackage[normalem]{ulem}   
\usepackage{enumitem}         

\usepackage[frozencache,cachedir=.]{minted} 

\usepackage{url}              
\usepackage{hyperref}         

\usepackage{array}

\usepackage{graphicx}
\usepackage{pifont}

\title{AsymTalker: Identity-Consistent Long-Term Talking Head Generation via Asymmetric Distillation}

\author{%
  Yuxin Lu\thanks{Equal contribution.} \\
  Soochow University \\
  \And
  Jiayang Sun\footnotemark[1] \\
  Soochow University \\
  \AND
  Guibo Zhu \\
  Institute of Automation, Chinese Academy of Sciences \\
  \And
  Min Cao\thanks{Corresponding author.} \\
  Soochow University \\
}

\begin{document}

\maketitle

\begin{abstract}
  Diffusion-based talking head generation has achieved remarkable visual quality, yet scaling it to long-term videos remains challenging. The widely adopted chunk-wise paradigm introduces two fundamental failures: (1) \emph{temporal-spatial misalignment} between static identity references and dynamic audio streams, and (2) cascading \emph{identity drift} propagated through self-generated continuity references across chunks. 
  To address both issues, we propose AsymTalker, a novel diffusion-based talking head generation method comprising Temporal Reference Encoding (TRE) and Asymmetric Knowledge Distillation (AKD). 
  First, TRE mitigates temporal-spatial misalignment by transforming the static identity image into a temporally coherent latent representation through encoding of a temporally replicated pseudo-video, without introducing additional parameters. 
  Second, AKD resolves the inherent conditioning dilemma in chunk-wise training: using ground-truth references causes train-inference mismatch, while self-generated references entangle supervision with identity drift. Our asymmetric design circumvents this by anchoring the teacher model with ground-truth continuity references to provide drift-free, chunk-level supervision, thereby avoiding the teacher bottleneck. Meanwhile, the student model learns under inference-aligned conditions, conditioned only on self-generated references, and is trained via distribution matching to preserve identity over long horizons.
  Extensive experiments show AsymTalker achieves state-of-the-art results on HDTF and VFHQ. It guarantees high-fidelity, identity-consistent synthesis over 600-second videos and reaches a real-time inference speed of 66 FPS. 

\end{abstract}

\section{Introduction}
\label{sec:introduction}

\begin{figure}[!t]
    \centering
    \includegraphics[width=1.0\linewidth]{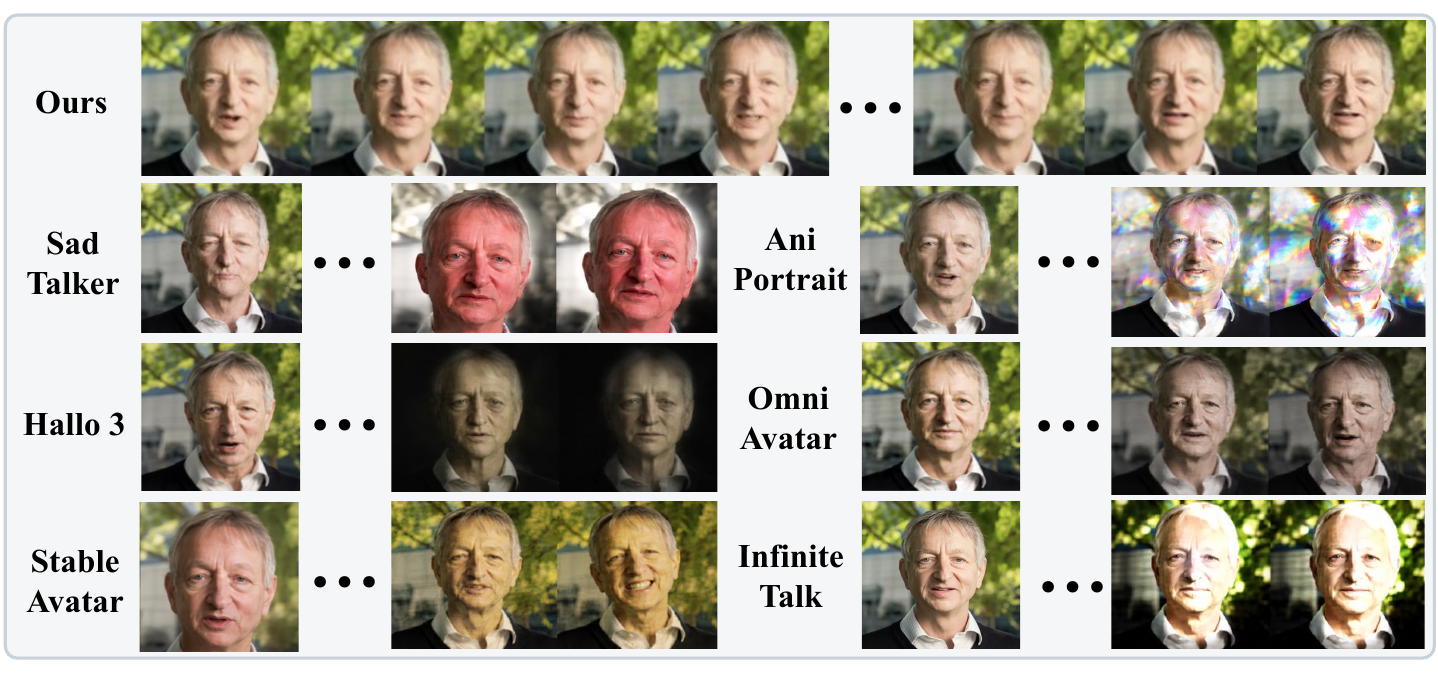}
    \caption{Long-term talking head generation quality comparison of our proposed AsymTalker and SOTA methods~\cite{zhang2023sadtalker,wei2024aniportrait,cui2025hallo3,gan2025omniavatar,tu2025stableavatar,yang2025infinitetalk}. Extensive samples are shown in Appendix~\ref{app:qualitative} and \url{https://asymtalker.com}. }
    \label{fig:show}
\end{figure}

The increasing demand for digital human interaction in virtual reality~\cite{chen2026rsatalker}, content creation~\cite{nazarieh2025magic}, and telepresence~\cite{ki2026avatar} has positioned realistic talking head generation as a pivotal research area. Talking head generation~\cite{zhang2023sadtalker} aims to synthesize photorealistic videos driven by a single identity image and audio stream. While traditional GAN-based frameworks~\cite{zhang2023sadtalker,liang2024wav2lip} have made significant strides in lip synchronization, they often fall short in rendering high-fidelity textures of the synthetic video. The recent integration of diffusion models~\cite{jiang2024loopy,tian2025emo2,fei2025skyreels,shen2025soulx,gan2025omniavatar} has largely overcome the visual quality limitation, setting a new standard for realism. However, scaling these diffusion-based frameworks to \textit{long-term} talking head generation remains a tremendous challenge due to prohibitive computational costs and memory constraints. To mitigate these, chunk-wise generation has been widely adopted into diffusion-based frameworks as a pragmatic scheme. 

Specifically, these methods typically generate each individual chunk on two distinct conditioning signals beyond the driving audio: an identity reference (i.e., the given static image) to preserve long-range identity consistency, and a continuity reference (i.e., the terminal frames of the preceding chunk) to maintain short-term coherence between adjacent chunks. Nevertheless, relying on these dual signals introduces critical weaknesses. 
(1) The identity reference provides solely static spatial cues, lacking temporal dynamics. In contrast, the driving audio dictates continuous frame variations in lip movements and facial expressions. Requiring the model to simultaneously condition on the static identity reference and the dynamic audio stream creates a severe temporal-spatial discrepancy, inherently degrading generation stability. 
(2) Constructing the continuity reference directly from the previously synthesized chunk inevitably leads to cascading errors. Any artifacts or deviations present in the prior continuity reference are directly inherited by the subsequent chunk, manifesting as catastrophic identity degradation over long-term generation---a limitation particularly acute in the talking head setting where identity is a visually salient anchor, as shown in Figure~\ref{fig:show}. 

To address the aforementioned limitations, we propose AsymTalker, a novel diffusion-based talking head generation method that achieves high-quality, identity-consistent long-term generation while maintaining real-time inference efficiency. Our method features two key innovations: 
(1) \textbf{Temporal Reference Encoding (TRE)}. To overcome the static nature of the identity reference, we temporally replicate the source image to match the duration of the target video chunk. By encoding this expanded sequence via a pre-trained 3D Variational Autoencoder (3D-VAE), we extract temporally coherent latent representations, naturally bridging the temporal misalignment between the static identity reference and the dynamic driving audio. 
(2) \textbf{Asymmetric Knowledge Distillation (AKD)}. While long-term identity drift caused by cascading continuity errors is widely recognized in diffusion-based video generation~\cite{huang2025self,cui2025self}, it remains fundamentally unresolved due to an inescapable conditioning dilemma when training: ground-truth references provide drift-free supervision but cause severe train-inference mismatch, whereas self-generated references align with inference but infect the supervision signal with compounding drift. We break this deadlock by decoupling these constraints across two specialized models: a full-step teacher, anchored by ground-truth references, provides drift-free supervision; meanwhile, a few-step student, driven by self-generated references, mirrors inference-time conditions and learns from the teacher via distribution matching distillation. Notably, as the teacher operates strictly at the chunk level with ground-truth anchoring, it requires no inherent long-term synthesis capacity. This structural advantage bypasses the teacher-bottleneck where short-horizon teachers upper-bound student performance~\cite{cui2025self}, enabling identity-consistent long-term synthesis. The few-step student further yields real-time inference as an additional benefit. 

Our primary contributions are summarized as follows:

\begin{itemize}[noitemsep, topsep=0pt, left=0pt]
\item We propose \textbf{Temporal Reference Encoding (TRE)} that mitigates temporal-spatial misalignment by temporally replicating the identity image and encoding it with a pre-trained 3D-VAE, yielding coherent conditions without extra parameters. 
\item We introduce \textbf{Asymmetric Knowledge Distillation (AKD)}, a teacher-student framework that decouples ground-truth supervision from self-generated conditioning, eliminating long-term identity drift while enabling real-time, chunk-wise inference. 
\item Experiments demonstrate that AsymTalker achieves identity-consistent long-term talking head generation over 600-second videos at 66 FPS real-time inference speed. 
\end{itemize}

\section{Related Work}

\paragraph{Talking Head Generation}
Early GAN-based methods, such as Wav2Lip~\cite{liang2024wav2lip} and SadTalker~\cite{zhang2023sadtalker}, improved lip-synchronization accuracy but frequently struggled with texture artifacts and limited expression vividness due to inherent GAN limitations, including training instability and mode collapse. To surpass these limitations, recent research has focused on diffusion-based frameworks, prized for their promising generative stability and texture fidelity. Within this paradigm, existing methods can be categorized by their identity reference injection mechanisms. Specifically, UNet-based architectures~\cite{tian2024emo,xu2024hallo,cui2024hallo2,chen2025echomimic,jiang2024loopy} typically incorporate an auxiliary Reference-UNet to inject identity references via intermediate layers. In contrast, DiT-based models focus on integrating identity cues through CLIP-based embeddings~\cite{tian2025emo2,fei2025skyreels, shen2025soulx}, visual token projection~\cite{qiu2025skyreels,deng2025avatarsync}, and Hunyuan-style (encode-then-repeat)~\cite{kong2024hunyuanvideo} latent conditioning~\cite{zhang2024letstalk,gan2025omniavatar} that first encodes the reference image via a VAE before repeating it along the temporal dimension. Despite their effectiveness, existing methods suffer from a fundamental temporal-spatial mismatch between static identity references and dynamic audio. Moreover, relying on chunk-wise paradigm for long-term generation inevitably propagates identity drift and struggles to support real-time synthesis. 


\paragraph{Diffusion Distillation}
Knowledge distillation aims to transfer the behavior of a powerful teacher model to a lightweight student model~\cite{hinton2015distilling}. This idea has been widely adopted to accelerate diffusion models, whose iterative denoising process usually requires many sampling steps~\cite{ho2020denoising}. Progressive distillation~\cite{salimans2022progressive} reduces sampling steps by repeatedly distilling a multi-step sampler into a shorter one, while consistency-based methods~\cite{song2023consistency,luo2023lcm} learn direct mappings along diffusion trajectories for few-step generation. More recently, Distribution Matching Distillation (DMD)~\cite{yin2024one,yin2024improved} matches the generated distribution of a student model to that of a teacher model, achieving high-quality one-step or few-step synthesis.
For video generation, recent works further explore distillation under temporal and causal constraints. CausVid~\cite{yin2025slow} distills a slow bidirectional video diffusion model into a faster autoregressive student, and Self Forcing~\cite{huang2025self} reduces the train-test gap by training on self-generated context. 
However, these methods primarily focus on accelerating general video synthesis. 
Notably, Self Forcing's context-agnostic teacher entangles supervision with drift-corrupted context, making it ill-suited for long-term talking head generation. In contrast, our AKD introduces an \emph{asymmetric} teacher-student conditioning paradigm, decouples drift-free supervision from inference-time alignment, explicitly targeting cross-chunk identity drift. 


\section{Method}

\subsection{Preliminaries}





\paragraph{Flow-matching-based Diffusion Models.}
Flow Matching (FM)~\cite{lipman2022flow,esser2024scaling} transforms a simple prior $p_0$ (e.g., Gaussian noise) into a complex data distribution $p_1$ by learning a velocity field $d\mathbf{x}_t = \bm{v}_\theta(t, \mathbf{x}_t) dt$. Using optimal transport with linear interpolation $\mathbf{x}_t=(1-t)\mathbf{x}_0 + t\mathbf{x}_1$, the model is trained to predict the target direction: 
\begin{equation}
    \mathcal{L}_{\text{FM}} = \mathbb{E}_{t, \mathbf{x}_0, \mathbf{x}_1} \left[ \| \bm{v}_\theta(t, \mathbf{x}_t) - (\mathbf{x}_1 - \mathbf{x}_0) \|^2_2 \right].
    \label{eq:fm_loss}
\end{equation}
During inference, samples are generated by solving the ODE from $t=0$ to $t=1$ using numerical solvers.

\paragraph{Backbone Architecture.}
We build upon the Wan2.1~\cite{wan2025wan} framework as our core backbone, which comprises $N$ stacked Diffusion Transformer (DiT) blocks and a well-trained 3D causal VAE consisting of an encoder $\mathcal{E}(\cdot)$ and a decoder $\mathcal{D}(\cdot)$. 
While the original Wan2.1 architecture is conditioned on text prompts and reference images, we adapt it for audio-driven generation by substituting the text encoder with a pre-trained Wav2Vec~\cite{baevski2020wav2vec} model to process audio conditions. 
Given a ground-truth video $\mathbf{V}\in\mathbb{R}^{T\times H\times W\times 3}$ containing $T$ frames at a resolution of $H\times W$, the 3D-VAE first compresses it into a latent representation $\mathbf{x}=\mathcal{E}(\mathbf{V})\in\mathbb{R}^{L\times h\times w\times C}$, where $L$, $h\times w$, and $C$ denote the latent temporal length, spatial resolution, and channel dimension, respectively. 
Building upon the standard FM objective defined in Eq.~(\ref{eq:fm_loss}), we formulate our training loss as: 
\begin{equation}
    \mathcal{L}_\text{diff}=\mathbb{E}_{t, \mathbf{x}_0, \mathbf{x}_1, \mathbf{c}_a, \mathbf{c}_I} \left[ \|\bm{v}_\theta(t, \mathbf{x}_t, \mathbf{c}_a, \mathbf{c}_I) - (\mathbf{x}_1-\mathbf{x}_0)\|^2_2 \right],
    \label{eq:diffusion-loss}
\end{equation}
where the target $\mathbf{x}_1$ corresponds to the encoded latent sequence $\mathbf{x}$. Here, $\mathbf{c}_a$ represents the audio condition extracted via Wav2Vec, and $\mathbf{c}_I$ denotes the identity reference. 

\subsection{Overall Architecture}
\label{sec:framework}

\begin{figure}[!t]
    \centering
    \includegraphics[width=1\linewidth]{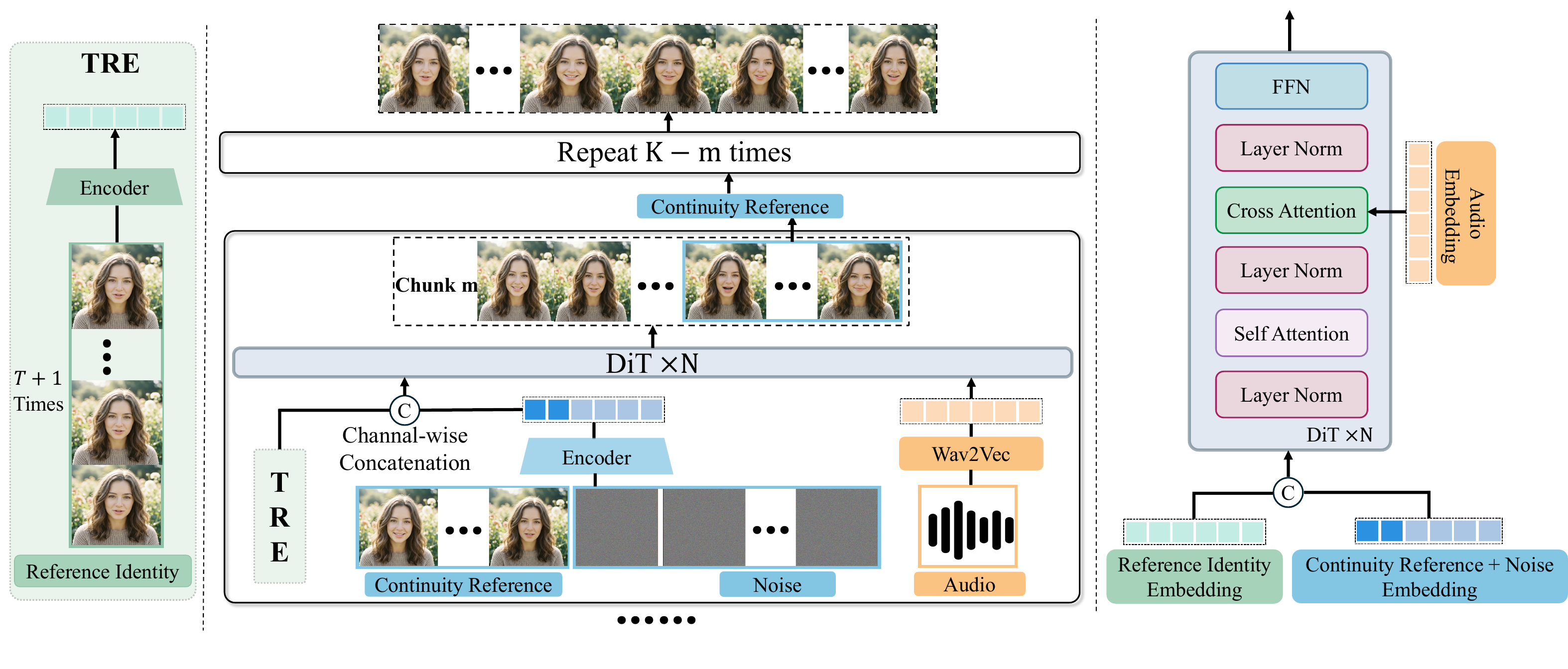}
    \caption{Overall architecture of our proposed AsymTalker. }
    \label{fig:framework}
\end{figure}

We build upon the chunk-wise paradigm for long-term generation (Figure~\ref{fig:framework}). Formally, the generated video is constructed by temporally concatenating $K$ causally consistent video chunks: $\hat{\mathbf{V}} = \{ \hat{\mathbf{V}}^{(1)}, \hat{\mathbf{V}}^{(2)}, \dots, \hat{\mathbf{V}}^{(K)} \}$, where each chunk $\hat{\mathbf{V}}^{(k)}\in\mathbb{R}^{T\times H\times W\times 3}$ is decoded via the 3D-VAE decoder $\mathcal{D}(\cdot)$ from the corresponding generated latent chunk $\hat{\mathbf{x}}^{(k)}\in\mathbb{R}^{L\times h\times w\times C}$. Each chunk is driven by an audio signal $\mathbf{c}_a^{(k)}$, extracted from the corresponding audio segment via the frozen Wav2Vec encoder followed by an MLP projection. Beyond this driving signal, two conditioning signals are required to ensure visual consistency, covering both intra-chunk identity preservation and inter-chunk temporal continuity: (1) \textbf{Identity reference} $\mathbf{c}_I$, encoding the source image to maintain long-range identity consistency. (2) \textbf{Continuity reference} $\bm{\kappa}^{(k)}$, carrying motion context from the $(k{-}1)$-th chunk via a decode-then-re-encode process: $\bm{\kappa}^{(k)}=\mathcal{E}\left(\mathcal{D}(\hat{\mathbf{x}}^{(k-1)}[-\tau:\,]) \right)$, where $\tau$ denotes the frame length of the reference window. Further explanation on the decode-then-re-encode process will be listed in Appendix~\ref{app:explanation-dtre}. 

Since $\mathbf{c}_a^{(k)}$ is inherently frame-aligned and temporally dynamic, it seamlessly guides the generation without requiring further structural adaptation. However, relying on the purely static $\mathbf{c}_I$ introduces severe temporal-spatial misalignment, while directly adopting $\bm{\kappa}^{(k)}$ inevitably leads to the accumulation of identity drift over time. We address these two issues with TRE (Sec.~\ref{sec:tre}) and AKD (Sec.~\ref{sec:akd}), respectively. 

\subsection{Temporal Reference Encoding}
\label{sec:tre}

A straightforward strategy to incorporate the identity reference is to encode the source image $I_{\mathrm{ref}}$ through a 3D-VAE and replicate the resulting latent along the temporal axis~\cite{zhang2024letstalk,gan2025omniavatar}, aiming to ensure every frame receives consistent identity conditioning. While the 3D-VAE natively supports single-frame inputs via spatial compression, processing a single image intrinsically bypasses its temporal modeling modules. Consequently, this \emph{encode-then-repeat} paradigm merely addresses the \emph{availability} of identity information. A fundamental issue remains unresolved: the extracted features are intrinsically \emph{temporally static}---each frame is conditioned on an identical copy of the same spatial features. When paired with a highly dynamic audio stream, this static spatial condition creates a severe temporal-spatial misalignment that inevitably degrades generation quality and stability. 

To bridge this gap, we propose TRE, which embeds temporal priors directly into the identity reference rather than solely focusing on feature availability. Instead of encoding first and replicating later, TRE reverses this sequence: we first temporally replicate $I_{\mathrm{ref}}$ to match the target chunk length $T$, constructing a pseudo-video $\{I_{\mathrm{ref}}\}_{T}$ consisting of pixel-wise identical frames. This sequence is then processed through the \emph{frozen} 3D-VAE encoder $\mathcal{E}$: 
\begin{equation}
    \mathbf{c}_I = \mathcal{E}(\{I_\mathrm{ref}\}_T)\in\mathbb{R}^{L\times h\times w\times C}.
    \label{eq:tre}
\end{equation}
Within $\mathcal{E}$, the 3D causal convolutions and temporal attention layers naturally process the input as a temporal sequence, embedding it into a learned spatiotemporal manifold. Critically, although the input frames are identical, the pretrained temporal priors within $\mathcal{E}$ infer a continuous temporal flow. This mechanism produces identity latents that carry rich spatial details while remaining structurally compatible with the dynamic audio condition. In effect, TRE transforms the identity reference from a \emph{static spatial anchor} into a \emph{temporally coherent condition}. Notably, this is achieved without introducing any learnable parameters or requiring per-frame supervision, and necessitates only a single encoding pass for all generated chunks. 

Finally, the resulting identity latents are fused with the initialized noisy input $\mathbf{x}_0^{(k)}$ via channel-wise concatenation: 
\begin{equation}
    \tilde{\mathbf{x}}_0^{(k)} =
    \mathrm{Concat}\bigl(\mathbf{x}_0^{(k)},\;
    \mathbf{c}_I\bigr) \in \mathbb{R}^{L\times h\times w\times 2C}.
    \label{eq:tre_concat}
\end{equation}

\subsection{Asymmetric Knowledge Distillation}
\label{sec:akd}

\begin{figure}[!t]
    \centering
    \includegraphics[width=1\linewidth]{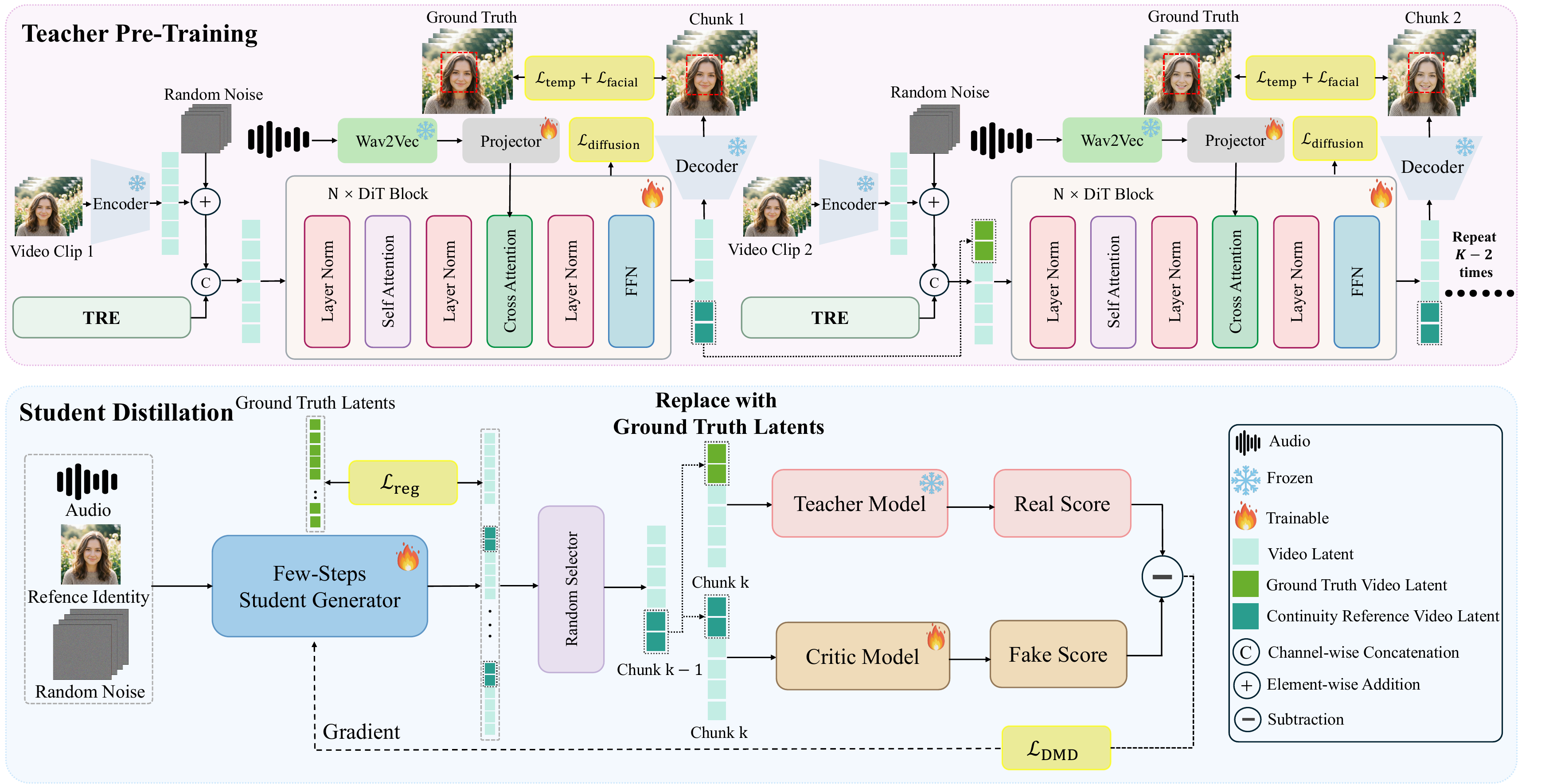}
    \caption{Training pipeline of AKD. The teacher $G_\mathrm{t}$ is conditioned on ground-truth continuity references $\bm{\kappa}_\text{gt}$ to provide drift-free supervision, while the student $G_\mathrm{s}$ and critic $G_\mathrm{c}$ are conditioned on self-generated references $\bm{\kappa}_\text{gen}$ to mirror inference-time conditions. }
    \label{fig:AKD}
\end{figure}

While TRE addresses the identity reference, the continuity reference $\bm{\kappa}^{(k)}$ introduces an equally critical challenge. Since $\bm{\kappa}^{(k)}$ is derived from \emph{previously generated} chunks, artifacts propagate forward as severe identity drift~\cite{huang2025self,cui2025self}. Standard training thus faces an inescapable dilemma: ground-truth references induce train-inference mismatch (\emph{exposure bias}~\cite{bengio2015scheduled}), while self-generated references entangle supervision with compounding drift. 

To resolve this fundamental bottleneck, we propose AKD, which decouples these conflicting constraints by deliberately splitting the conditioning inputs across the networks involved in distillation (Figure~\ref{fig:AKD}). Specifically, a teacher model $G_\mathrm{t}$ is conditioned on ground-truth references $\bm{\kappa}_\text{gt}^{(k)}$ to establish a drift-free supervision baseline. Concurrently, a student model $G_\mathrm{s}$ is conditioned on self-generated references $\bm{\kappa}_\text{gen}^{(k)}$ to perfectly mirror the inference-time environment. Through distribution matching, the student inherits the teacher's drift-free generation capabilities while operating under strictly aligned conditions. This asymmetric design naturally operates in two stages: teacher pre-training and student distillation.

\paragraph{Teacher Pre-training.}
The teacher model $G_\mathrm{t}$, initialized from the full-step DiT backbone, is pre-trained exclusively with ground-truth continuity references. Specifically, we extract $\bm{\kappa}_\text{gt}^{(k)} = \mathcal{E}(\mathcal{D}(\mathbf{x}_\text{gt}^{(k-1)}[-\tau:\,]))$ from the preceding ground-truth chunk. Conditioned on the dynamic audio $\mathbf{c}_a^{(k)}$, identity reference $\mathbf{c}_I$, and $\bm{\kappa}_\text{gt}^{(k)}$, the teacher predicts the velocity field for the target latent $\mathbf{x}_\text{gt}^{(k)}$. As $\bm{\kappa}_\text{gt}^{(k)}$ encapsulates a flawless, drift-free motion context, $G_\mathrm{t}$ establishes a robust supervision baseline that is fundamentally immune to the compounding exposure bias of chunk-wise inference. 

Beyond the standard flow-matching objective $\mathcal{L}_\text{diff}$ defined in Eq.~\ref{eq:diffusion-loss}, we further introduce two pixel-level auxiliary losses---temporal gradient consistency loss $\mathcal{L}_\text{temp}^{(k)}$ and facial region focalization loff $\mathcal{L}_\text{facial}^{(k)}$---to enhance temporal smoothness and facial fidelity. Formal definitions are shown in Appendix~\ref{app:teacher-loss}. 
The overall teacher pre-training objective is:
\begin{equation}
    \mathcal{L}_\mathrm{teacher} = \frac{1}{K}\sum_{k=1}^{K}\left( \mathcal{L}_\text{diff}^{(k)} +  \mathcal{L}_\text{temp}^{(k)} +  \mathcal{L}_\text{facial}^{(k)} \right),
    \label{eq:teacher_loss}
\end{equation}
where $\mathcal{L}_\text{diff}^{(k)}$ follows the standard FM objective in Eq.~(\ref{eq:diffusion-loss}) with $\bm{\kappa}_\text{gt}^{(k)}$ as an additional condition. Upon completion of this pre-training phase, the weights of the teacher model $G_\mathrm{t}$ are strictly frozen to serve as a stable target for the subsequent student distillation. 

\paragraph{Student Distillation.}
Initialized from the pre-trained teacher $G_\mathrm{t}$, the student generator $G_\mathrm{s}$ performs efficient few-step synthesis conditioned on \emph{self-generated} continuity references. Concretely, for chunk $k$, the student produces $\hat{\mathbf{x}}^{(k)} = G_\mathrm{s}(\tilde{\mathbf{x}}_0^{(k)} \,|\, \mathbf{c}_a^{(k)}, \mathbf{c}_I, \bm{\kappa}_\text{gen}^{(k)})$, where $\bm{\kappa}_\text{gen}^{(k)}$ is constructed from the student's own output of the preceding chunk via decode-then-re-encode. This conditioning exactly mirrors the inference-time environment, inherently avoiding exposure bias. 

We adopt DMD~\cite{yin2024one} as the alignment objective, with conditioning inputs deliberately split across the three networks involved. The teacher $G_\mathrm{t}$ is conditioned on $\bm{\kappa}_\text{gt}$, defining the drift-free target distribution $p_\text{real}$. The student $G_\mathrm{s}$ and an auxiliary critic $G_\mathrm{c}$ are both conditioned on $\bm{\kappa}_\text{gen}$, with the former generating samples under inference-aligned conditions and the latter tracking the resulting distribution $p_\text{fake}$. This asymmetric topology is the structural core of AKD. The student is trained to minimize the reverse KL divergence: 
\begin{equation}
    \mathcal{L}_\text{DMD} = D_\text{KL}(p_\text{fake} \,\|\, p_\text{real}),
    \label{eq:dmd}
\end{equation}
whose gradient is approximated via score distillation from the discrepancy between scores provided by $G_\mathrm{t}$ and $G_\mathrm{c}$~\cite{yin2024one}. 

While DMD effectively transfers the teacher's generation quality, relying solely on distribution matching can allow the student's generation trajectory to drift, particularly under our asymmetric conditioning. To strictly constrain this, we introduce a Huber-style~\cite{meyer2021alternative} \textbf{regression anchoring} loss. This loss stabilizes the generation trajectory without enforcing overly rigid path-level imitation: 
\begin{equation}
    \mathcal{L}_\text{reg}^{(k)} =
    \begin{cases}
    \frac{1}{2}\|\hat{\mathbf{x}}^{(k)} - \mathbf{x}_\text{gt}^{(k)}\|_2^2,
    & \|\hat{\mathbf{x}}^{(k)} - \mathbf{x}_\text{gt}^{(k)}\|_1 \le 1,\\
    \|\hat{\mathbf{x}}^{(k)} - \mathbf{x}_\text{gt}^{(k)}\|_1 - \frac{1}{2}, & \text{otherwise}.
    \end{cases}
    \label{eq:reg_loss}
\end{equation}

The overall student training objective is defined as:
\begin{equation}
    \mathcal{L}_\mathrm{student} = \frac{1}{K}\sum_{k=1}^{K}\left( \mathcal{L}_\text{DMD}^{(k)} + \lambda_\text{reg} \mathcal{L}_\text{reg}^{(k)} \right).
    \label{eq:student_loss}
\end{equation}

This asymmetric design narrows the teacher's required competency to chunk-level synthesis alone, structurally bypassing the teacher-bottleneck~\cite{cui2025self} and unlocking drift-free over arbitrarily long-term generation. The few-step student further yields real-time inference. 

\paragraph{Inference.}
At inference, the trained student $G_\mathrm{s}$ generates each chunk with $N=4$ denoising steps. 
After generating chunk $k$, the tail frames are decoded and re-encoded to produce $\bm{\kappa}_\text{gen}^{(k+1)} = \mathcal{E}(\mathcal{D}(\hat{\mathbf{x}}^{(k)}[-\tau:\,]))$, which is then fed into the next chunk.
This loop proceeds autoregressively for $K$ iterations, producing the complete long-term video.

\section{Experiments}
\label{sec:experiments}
\subsection{Experimental Settings}
\label{sec:experimental-settings}

\paragraph{Datasets and Baselines.}
We construct our training corpus by aggregating AVSpeech~\cite{ephrat2018looking}, HDTF~\cite{zhang2021flow}, OpenHumanVid~\cite{li2025openhumanvid}, TalkVid~\cite{chen2025talkvid}, VFHQ~\cite{xie2022vfhq}, alongside a self-collected dataset, yields 217 hours of high-quality, synchronized audio-visual pairs. Comprehensive details regarding the dataset construction and processing pipeline are provided in the Appendix~\ref{dataset}. 
For evaluation, we sample 100 test videos uniformly at random from both HDTF (emphasis on diverse facial expressions) and VFHQ (contains high-fidelity and temporally consistent visuals). All compared methods are evaluated using identical reference images and driving audio sequences. 
We benchmark AsymTalker against representative state-of-the-art (SOTA) models across GAN-based~\cite{zhang2023sadtalker}, UNet-based~\cite{wei2024aniportrait}, and DiT-based architectures~\cite{gan2025omniavatar,cui2025hallo3,tu2025stableavatar,yang2025infinitetalk}. To ensure fairness, we adopt the official implementations and recommended inference configurations for all baselines.

\paragraph{Evaluation Metrics.}
We evaluate visual quality and temporal consistency with Fréchet Inception Distance (FID)~\cite{heusel2017gans} and Fréchet Video Distance (FVD)~\cite{unterthiner2018towards}, respectively, where lower values are better. Lip synchronization is measured by SyncNet~\cite{chung2016out}, using Synchronization Confidence (Sync-C) and Synchronization Distance (Sync-D); higher Sync-C and lower Sync-D indicate better audio-visual alignment. 

\subsection{Results}

\subsubsection{Quantitative Results}

\begin{table}[!t]
\caption{Comparison with SOTA methods on HDTF and VFHQ. The best result is marked in \textbf{bold} and the second-best result is \underline{underlined}. }
\label{tab:sota_main}
\vspace{0.5em}
\centering
\scalebox{1}{\setlength{\tabcolsep}{5mm}
\small\begin{tabular}{llcccc}
\toprule
Dataset & Method & FID$\downarrow$ & FVD$\downarrow$ & Sync-C$\uparrow$ & Sync-D$\downarrow$ \\
\cmidrule{1-6}
\multirow{7}{*}{HDTF}
& SadTalker~\cite{zhang2023sadtalker}~(\citeyear{zhang2023sadtalker})        & 21.96 & 205.77 & 6.24 & \underline{8.37} \\
& AniPortrait~\cite{wei2024aniportrait}~(\citeyear{wei2024aniportrait})      & 21.33 & 238.48 & 2.97 & 11.91 \\
& OmniAvatar~\cite{gan2025omniavatar}~(\citeyear{gan2025omniavatar})        & \textbf{12.23} & 155.71 & 3.89 & 10.11 \\
& Hallo3~\cite{cui2025hallo3}~(\citeyear{cui2025hallo3})                 & 14.75 & \underline{134.94} & 4.21 & 10.01 \\
& StableAvatar~\cite{tu2025stableavatar}~(\citeyear{tu2025stableavatar})     & 15.89 & 146.79 & \underline{7.01} & 8.49 \\
& InfiniteTalk~\cite{yang2025infinitetalk}~(\citeyear{yang2025infinitetalk})   & 16.01 & 163.22 & 6.41 & 8.50 \\
& \textbf{AsymTalker (Ours)}                      & \underline{13.72} & \textbf{116.78} & \textbf{8.11} & \textbf{7.25} \\
\cmidrule{1-6}
\multirow{7}{*}{VFHQ}
& SadTalker~\cite{zhang2023sadtalker}~(\citeyear{zhang2023sadtalker})        & 45.56 & 301.89 & \underline{6.05} & 9.15 \\
& AniPortrait~\cite{wei2024aniportrait}~(\citeyear{wei2024aniportrait})      & 50.22 & 288.50 & 2.74 & 11.97 \\
& OmniAvatar~\cite{gan2025omniavatar}~(\citeyear{gan2025omniavatar})        & 32.91 & 267.52 & 3.52 & 11.64 \\
& Hallo3~\cite{cui2025hallo3}~(\citeyear{cui2025hallo3})                 & 38.70 & \underline{192.06} & 4.88 & 9.76 \\
& StableAvatar~\cite{tu2025stableavatar}~(\citeyear{tu2025stableavatar})     & \underline{31.55} & 249.40 & 6.04 & 9.32 \\
& InfiniteTalk~\cite{yang2025infinitetalk}~(\citeyear{yang2025infinitetalk})   & 33.42 & 201.52 & 5.99 & \underline{9.06} \\
& \textbf{AsymTalker (Ours)}                      & \textbf{23.25} & \textbf{182.35} & \textbf{6.41} & \textbf{8.50} \\
\bottomrule
\end{tabular}}
\end{table}

\paragraph{Comparison with SOTA methods.}
Table~\ref{tab:sota_main} reports the main comparison on HDTF and VFHQ. On HDTF, AsymTalker achieves the best FVD (116.78), Sync-C (8.11), and Sync-D (7.25), and ranks second on FID. 
The leading FVD and Sync-C indicates that TRE's temporally coherent identity condition frees the audio-driven motion stream to realize expressive dynamics without sacrificing temporal coherence, aligning with HDTF's emphasis on diverse expressions. 
On VFHQ, AsymTalker outperforms all baselines across all four metrics. Compared to the second-best results, it reduces FID from 31.55 to 23.25 and FVD from 192.06 to 182.35. 
Notably, the VFHQ dataset, characterized by high-fidelity imagery and strong temporal consistency, provides an ideal stress test for both image-level realism and video-level coherence, aligning closely with the strengths of our AsymTalker. 
While these short-clip benchmarks already establish AsymTalker as SOTA, its more substantial advantages, namely \emph{identity-consistent long-term video generation} and \emph{real-time inference capability}, are further demonstrated in the long-term and efficiency analyses below. 

\begin{table}[!t]
\centering
\begin{minipage}{0.48\linewidth}
    \centering
    \vspace{-1.3em}
    \caption{Ablation study of the proposed TRE and AKD modules. The baseline configuration utilizes CLIP-based identity injection and relies solely on the teacher model for generation. 
    }
    \label{tab:ablation_main}
    \vspace{0.5em}
    \scalebox{0.95}{\setlength{\tabcolsep}{1.5mm}
    \small\begin{tabular}{cccccc}
    \toprule
    \multicolumn{2}{c}{Modules} & \multicolumn{4}{c}{Metrics} \\
    \cmidrule(lr){1-2} \cmidrule(lr){3-6}
    TRE & AKD & FID$\downarrow$ & FVD$\downarrow$ & Sync-C$\uparrow$ & Sync-D$\downarrow$ \\
    \midrule
    $\times$ & $\times$ & 19.92 & 232.50 & 6.52 & 10.43 \\
    $\checkmark$ & $\times$ & \underline{15.26} & \underline{148.15} & 7.61 & 9.33 \\
    $\times$ & $\checkmark$ & 16.71 & 150.23 & \textbf{8.19} & \underline{8.05} \\
    $\checkmark$ & $\checkmark$ & \textbf{13.72} & \textbf{116.78} & \underline{8.11} & \textbf{7.25} \\
    \bottomrule
    \end{tabular}}
\end{minipage}\hspace{1em}%
\begin{minipage}{0.48\linewidth}
    \centering
    \caption{Ablation study on TRE, where CLIP, VTP and ETR represent CLIP-based embedding, visual token projection and Encode-Then-Repeat paradigm, respectively. }
    \label{tab:ablation_tre}
    \vspace{0.5em}
    \scalebox{0.95}{\setlength{\tabcolsep}{1.2mm}
    \small\begin{tabular}{ccccc}
    \toprule
    Setting & FID$\downarrow$ & FVD$\downarrow$ & Sync-C$\uparrow$ & Sync-D$\downarrow$ \\
    \midrule
    CLIP       & \underline{16.71} & \underline{150.23} & \textbf{8.19} & \underline{8.05} \\
    VTP    & 18.99 & 201.13 & 4.24 & 10.89 \\
    ETR & 17.45 & 167.89 & 5.67 & 9.12 \\
    \textbf{TRE (Ours)}        & \textbf{13.72} & \textbf{116.78} & \underline{8.11} & \textbf{7.25} \\
    \bottomrule
    \end{tabular}}
\end{minipage}

\end{table}

\begin{figure}[!t]
    \centering
    \begin{minipage}{0.48\linewidth}
        \centering
        \includegraphics[width=\linewidth]{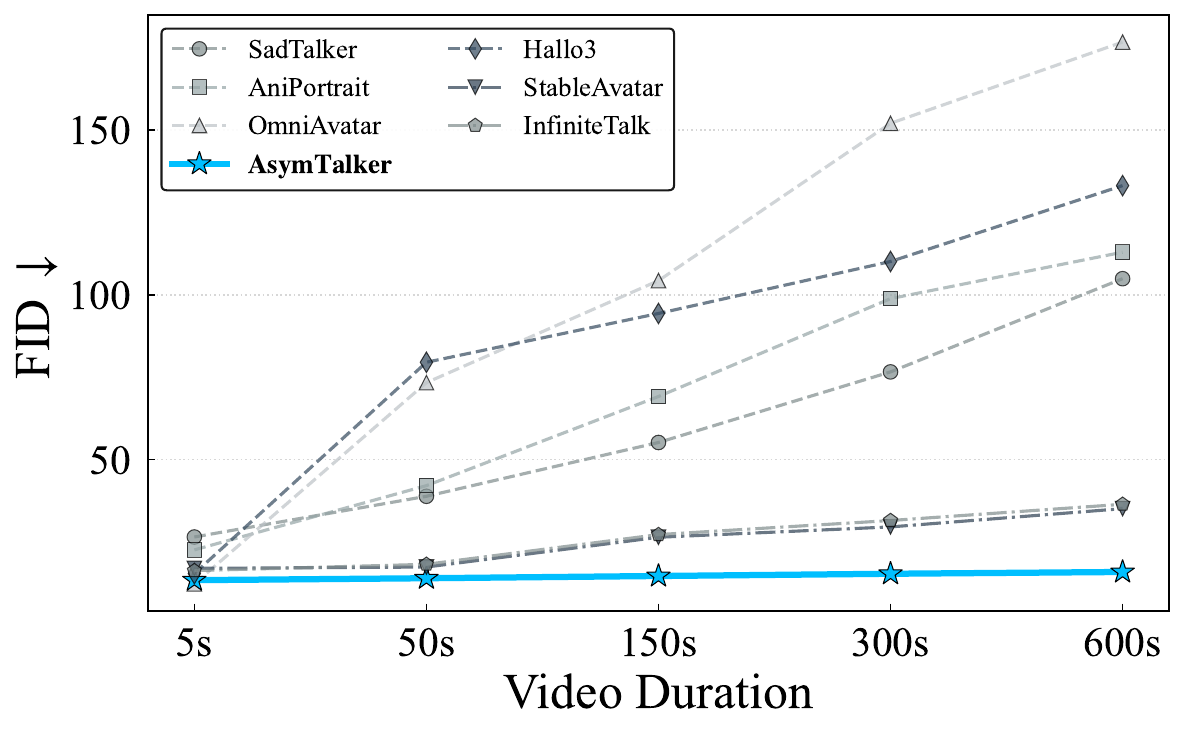}
        \caption{FID comparison over different video durations. }
        \label{fig:long-term}
    \end{minipage}\hspace{0.8em}%
    \begin{minipage}{0.48\linewidth}
        \centering
        \includegraphics[width=\linewidth]{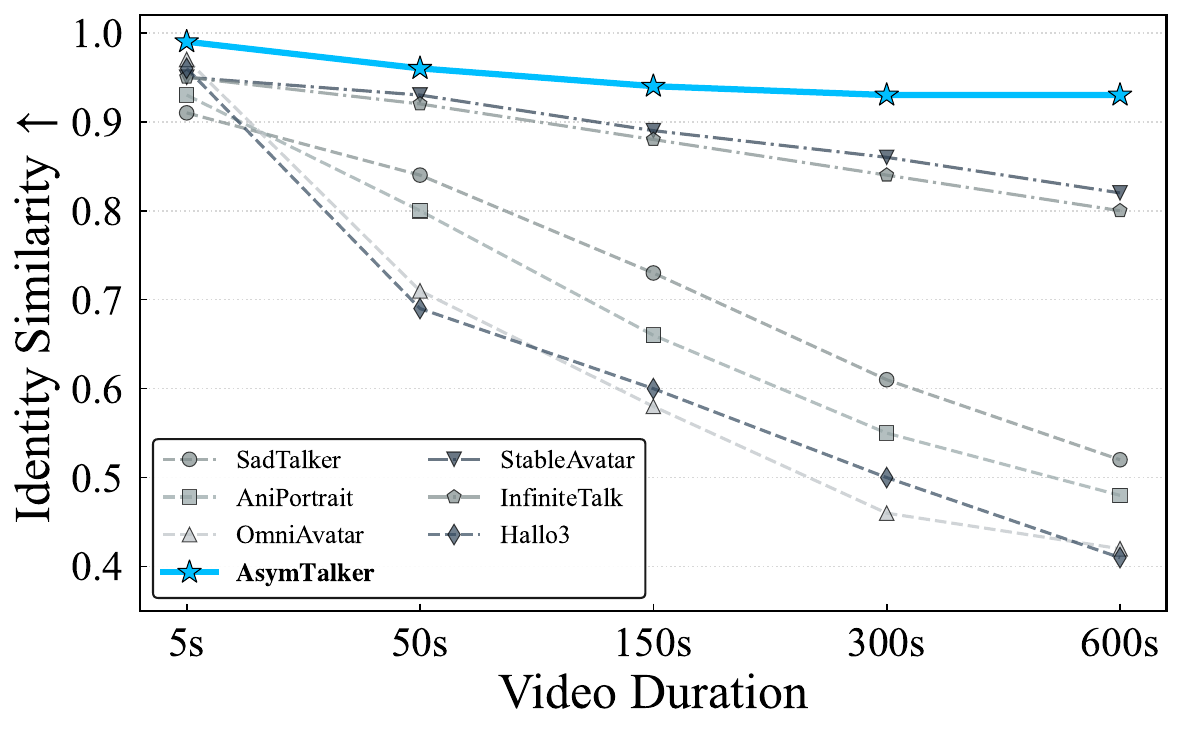}
        \caption{Identity similarity comparison over different video durations based on ArcFace~\cite{deng2022arcface}. }
        \label{fig:id-sim}
    \end{minipage}
\end{figure}

\paragraph{Long-term Identity Stability.}
Figures~\ref{fig:long-term} and~\ref{fig:id-sim} jointly evaluate long-term identity stability by generating continuous videos up to 600 seconds and reporting FID and ArcFace identity similarity~\cite{deng2022arcface} at different timestamps. Most baselines exhibit clear monotonic degradation as duration grows. For example, OmniAvatar~\cite{gan2025omniavatar}'s FID rises from 12.27 at 5s to 176.75 at 600s, while SadTalker~\cite{zhang2023sadtalker}, AniPortrait~\cite{wei2024aniportrait}, and Hallo3~\cite{cui2025hallo3} also degrade substantially over long horizons. StableAvatar~\cite{tu2025stableavatar} and InfiniteTalk~\cite{yang2025infinitetalk} are more robust on FID, yet their ArcFace identity similarity still drops markedly, suggesting that low FID alone does not imply preserved identity. By contrast, AsymTalker remains nearly flat on both metrics, with FID rising only from 13.41 to 15.89 across 600 seconds and ArcFace similarity staying within a narrow band. 
This directly validates AKD's identity-consistent property by structurally bypassing the teacher-bottleneck (Sec.~\ref{sec:akd}). 

\paragraph{Inference Efficiency.}
\begin{wrapfigure}[10]{r}{0.45\linewidth}
\vspace{-2em}
\centering
\includegraphics[width=\linewidth]{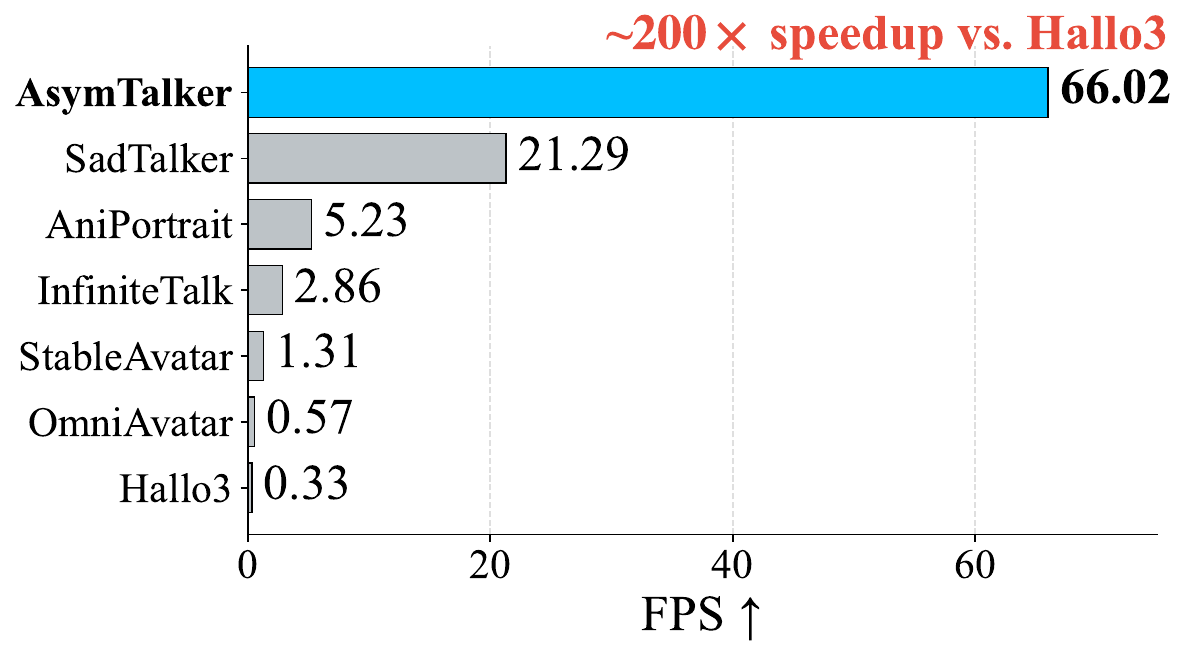}
\caption{Inference speed (FPS) comparison with SOTA methods.}
\label{fig:fps}
\end{wrapfigure}
Figure~\ref{fig:fps} reports inference throughput on 30-second $512 \times 512$ videos, measured under identical conditions on an A100 GPU. AsymTalker achieves 66.02 FPS, surpassing the 25 FPS threshold for real-time performance, despite employing no specialized acceleration techniques. This represents about $3\times$ speedup over SadTalker, the fastest baseline in our comparison, and about $200\times$ improvement over Hallo3. 

\subsubsection{Qualitative Results}


The qualitative comparison in Figure~\ref{fig:show} evaluates visual quality under the same input of the reference image and driving audio. 
AsymTalker generates high-fidelity talking-head videos characterized by clear facial textures, natural mouth shapes, and expressive coherent facial dynamics. 
In contrast, SadTalker, StableAvatar, and AniPortrait suffer from unnatural head motion or static expressions, while OmniAvatar and Hallo3 still exhibit identity drift and background instability over long-term generation. 
Overall, AsymTalker consistently preserves identity while generating natural expressions and stable visual coherence over extended sequences. Additional qualitative results are provided in the Appendix~\ref{app:qualitative}, with richer demos available at \url{https://asymtalker.com}.

\subsection{Ablation Study}

We conduct ablations on HDTF dataset to isolate the contribution of the two core designs: TRE for temporally compatible identity conditioning and AKD for drift-free distillation. Unless otherwise stated, all variants use the same backbone, training data, resolution, chunk length, optimizer, and evaluation protocol as the full model.

Table~\ref{tab:ablation_main} studies the complementarity of TRE and AKD. Removing both components gives the worst performance, with FVD rising to 232.50 and Sync-D increasing to 10.43, which confirms that the plain chunk-wise diffusion-distillation baseline is vulnerable to both temporal condition mismatch and accumulated continuity errors. Adding TRE alone substantially improves visual quality and temporal coherence, reducing FID from 19.92 to 15.26 and FVD from 232.50 to 148.15. 
This gain stems from TRE ability to enrich the identity reference beyond a static spatial prior, enabling temporally aware conditioning. 
Adding AKD alone improves both image-level realism and lip synchronization on these short clips, reaching the best isolated Sync-C of 8.19. 
When combined, TRE and AKD—forming our full AsymTalker—deliver the strongest overall performance, demonstrating that they jointly address two distinct failure modes: temporal-spatial misalignment (mitigated by TRE) and cascading identity drift (suppressed by AKD). 

\paragraph{Effect of TRE.}
\label{sec:ablation-tre}

Table~\ref{tab:ablation_tre} compares TRE against three representative identity injection strategies under identical experimental conditions: \textit{CLIP-based Embedding}~\cite{fei2025skyreels}, \textit{Visual Token Projection}~\cite{deng2025avatarsync}, and \textit{Encode-Then-Repeat}~\cite{zhang2024letstalk,gan2025omniavatar} (see Appendix~\ref{app:identity-strategies} for implementation details). 
TRE consistently achieves superior performance in FID, FVD, and Sync-D. It first repeats the reference image in pixel space to form a pseudo-video and subsequently encodes it with the frozen 3D-VAE. This design crucially leverages the VAE's learned temporal priors to refine the identity representation before fusion with the audio-driven motion stream, yielding an identity condition that is both spatially faithful and temporally coherent.

\paragraph{Effect of AKD.}


\begin{table}[!t]
\centering
\caption{Ablation study on AKD. ``Gen.'' and ``GT'' denote using self-generated and ground-truth continuity references, respectively. }
\label{tab:ablation_akd}
\vspace{0.5em}
\small\begin{tabular}{lcccc}
\toprule
Setting (Teacher / Student) & FID$\downarrow$ & FVD$\downarrow$ & Sync-C$\uparrow$ & Sync-D$\downarrow$ \\
\midrule
Symmetric GT (GT / GT)      & 22.73 & 226.41 & 4.90 & 10.88 \\
Symmetric Gen. (Gen. / Gen.) & 16.05 & 165.24 & 5.89 & 10.12 \\
\textbf{Asymmetric (GT / Gen.)} & \textbf{13.72} & \textbf{116.78} & \textbf{8.11} & \textbf{7.25} \\
\bottomrule
\end{tabular}
\end{table}

Table~\ref{tab:ablation_akd} ablates the asymmetric conditioning strategy in AKD. When both teacher and student rely on self-generated continuity references, the student remains aligned with inference; however, the teacher-side supervision signal suffers from accumulated errors, degrading performance. Alternatively, when both networks use ground-truth references, the supervision signal is clean, but the student is trained under a condition that it cannot access during deployment, causing a severe train-inference mismatch and the weakest results. This debunks the naive assumption that a drift-free teacher yields a better student, reinforcing our claim in Sec.~\ref{sec:akd} that bypassing the teacher-bottleneck demands structural decoupling over symmetric supervision. Finally, our AKD assigns these two roles asymmetrically: the teacher is conditioned on ground-truth references to define a drift-free target distribution, while the student is conditioned on self-generated references to mirror the inference-time environment. This design achieves the best results.

\section{Conclusion}
We propose AsymTalker, which resolves two structural weaknesses of chunk-wise diffusion-based talking head generation: TRE bridges temporal-spatial misalignment by processing frame-replicated pseudo-videos through a frozen 3D-VAE; AKD decouples the conflicting condition demands of chunk-wise training across a teacher-student pair, structurally bypassing the teacher-bottleneck to achieve identity-consistent long-term synthesis. Extensive experiments confirm competitive performance on the HDTF and VFHQ datasets, identity stability over 600-second videos, and 66 FPS real-time inference. 

\bibliographystyle{plainnat}
\bibliography{references}

\newpage
\appendix

\section*{Appendix}

\section{Limitations}
\label{app:limitations}

Although AsymTalker achieves strong identity consistency and long-term stability, its generated head motion is slightly less dynamic than some existing methods. We believe this may be partly caused by our emphasis on suppressing cross-chunk identity drift and preserving stable continuity references, which can make the model favor conservative motion patterns during long-term generation. In addition, the student distillation stage may further smooth high-amplitude motion in order to maintain stable few-step inference. In future work, we plan to improve motion expressiveness by introducing stronger pose-aware supervision, motion diversity regularization, and adaptive control over the trade-off between identity stability and head-motion amplitude.

\section{Implementation Details}
\label{app:implementation-details}
All experiments are conducted on 16 NVIDIA A100 GPUs. Following Wan2.1~\cite{wan2025wan}, all video samples are resized to $512 \times 512$ and processed as 81-frame chunks at 25 FPS. We set the continuity reference window to $\tau=3$ and the regression anchoring weight to $\lambda_\text{reg}=0.2$. The teacher model uses 1,000 denoising timesteps, while the student model is distilled to 4 denoising steps for efficient inference. We optimize the models with AdamW~\cite{loshchilov2017decoupled}, use a batch size of 4, and adopt bfloat16 mixed-precision training with Fully Sharded Data Parallel (FSDP)~\cite{zhao2023pytorch}. The teacher is first pre-trained for 15,000 steps and then fixed to supervise a 1,600-step student distillation stage. In our training environment, teacher pre-training takes about 5 days on 16 A100 GPUs, corresponding to approximately $1{,}920$ GPU-hours. Student distillation takes about 2 days on the same setup, corresponding to approximately $768$ GPU-hours. Unless otherwise specified, all ablation variants share the same data, backbone, resolution, chunk length, optimizer, and training schedule.

\section{Distribution Matching Distillation Details}
\label{app:dmd}

This section gives a self-contained description of Distribution Matching Distillation (DMD)~\cite{yin2024one,yin2024improved}. DMD distills a pretrained diffusion or flow model into a fast generator by directly matching the distribution induced by the student to the target distribution represented by the teacher. Unlike trajectory-level distillation methods that ask the student to imitate every intermediate denoising step, DMD optimizes the generated sample distribution itself through a reverse-KL objective estimated from score differences.

\paragraph{DMD setup.}
For a training item, let $\mathbf{c}^{(k)}$ denote the condition used for the $k$-th chunk, e.g., the audio condition, identity reference, and any continuity context used by the generator. Given Gaussian noise $\mathbf{z}^{(k)}\sim\mathcal{N}(\mathbf{0},\mathbf{I})$, the few-step student generator produces
\begin{equation}
    \hat{\mathbf{x}}_\mathrm{s}^{(k)}
    =
    G_\mathrm{s}\bigl(\mathbf{z}^{(k)} \mid \mathbf{c}^{(k)}\bigr),
    \label{eq:appendix_student_sample}
\end{equation}
which induces a generated distribution $p_\mathrm{s}(\mathbf{x}\mid\mathbf{c}^{(k)})$. The frozen teacher $G_\mathrm{t}$ represents the target distribution $p_\mathrm{t}(\mathbf{x}\mid\mathbf{c}^{(k)})$. Since the student distribution changes during optimization, DMD additionally trains a fake-score model $G_\mathrm{c}$ to estimate the score of the current student samples.

\paragraph{Reverse-KL objective.}
DMD minimizes the reverse KL divergence from the student distribution to the teacher distribution:
\begin{equation}
    \mathcal{L}_\text{DMD}^{(k)}
    =
    D_\text{KL}\left(
    p_\mathrm{s}(\mathbf{x}\mid\mathbf{c}^{(k)})
    \,\middle\|\,
    p_\mathrm{t}(\mathbf{x}\mid\mathbf{c}^{(k)})
    \right).
    \label{eq:appendix_reverse_kl}
\end{equation}
The reverse-KL form is suitable for distillation because samples are drawn from the student and then pushed toward regions where the teacher assigns high probability.

\paragraph{Score estimation under flow matching.}
For the flow-matching parameterization used by our backbone, we sample a noisy point along the linear path between Gaussian noise and the student output:
\begin{equation}
    \mathbf{x}_t^{(k)}
    =
    (1-t)\bm{\epsilon}^{(k)} + t\hat{\mathbf{x}}_\mathrm{s}^{(k)},
    \qquad
    t\sim\mathcal{U}(0,1),\quad
    \bm{\epsilon}^{(k)}\sim\mathcal{N}(\mathbf{0},\mathbf{I}).
    \label{eq:appendix_dmd_path}
\end{equation}
The velocity predictions of the teacher and fake-score model are converted into score estimates with the standard flow-matching score conversion operator $\mathcal{S}(\cdot)$:
\begin{equation}
    \mathbf{s}_\mathrm{t}^\text{real}
    =
    \mathcal{S}\left(
    \bm{v}_\mathrm{t}(t,\mathbf{x}_t^{(k)},\mathbf{c}^{(k)}),
    \mathbf{x}_t^{(k)},t
    \right),
    \quad
    \mathbf{s}_\mathrm{c}^\text{fake}
    =
    \mathcal{S}\left(
    \bm{v}_\mathrm{c}(t,\mathbf{x}_t^{(k)},\mathbf{c}^{(k)}),
    \mathbf{x}_t^{(k)},t
    \right).
    \label{eq:appendix_score_conversion}
\end{equation}
Here $\mathbf{s}_\mathrm{t}^\text{real}$ estimates the score of the teacher distribution and $\mathbf{s}_\mathrm{c}^\text{fake}$ estimates the score of the current student distribution. The student gradient can then be written as
\begin{equation}
    \nabla_{\theta_\mathrm{s}} \mathcal{L}_\text{DMD}^{(k)}
    =
    \mathbb{E}_{t,\bm{\epsilon}}
    \left[
    w(t)
    \left(
    \mathbf{s}_\mathrm{c}^\text{fake}
    -
    \mathbf{s}_\mathrm{t}^\text{real}
    \right)
    \frac{\partial \hat{\mathbf{x}}_\mathrm{s}^{(k)}}{\partial \theta_\mathrm{s}}
    \right],
    \label{eq:appendix_dmd_grad}
\end{equation}
where $w(t)$ is the timestep weighting. Intuitively, the score difference points the student samples away from regions over-represented by the current student distribution and toward regions favored by the teacher distribution.

\paragraph{Fake-score critic.}
The fake-score model $G_\mathrm{c}$ is trained on stop-gradient student samples so that it tracks the evolving generated distribution without updating the student through this auxiliary objective. With
\begin{equation}
    \tilde{\mathbf{x}}_t^{(k)}
    =
    (1-t)\bm{\epsilon}^{(k)}
    +
    t\,\mathrm{stopgrad}\left(\hat{\mathbf{x}}_\mathrm{s}^{(k)}\right),
    \label{eq:appendix_fake_path}
\end{equation}
the critic is optimized by the flow-matching loss
\begin{equation}
    \mathcal{L}_\text{critic}^{(k)}
    =
    \mathbb{E}_{t,\bm{\epsilon}}
    \left[
    \left\|
    \bm{v}_\mathrm{c}\left(t,\tilde{\mathbf{x}}_t^{(k)},\mathbf{c}^{(k)}\right)
    -
    \left(\mathrm{stopgrad}\left(\hat{\mathbf{x}}_\mathrm{s}^{(k)}\right)-\bm{\epsilon}^{(k)}\right)
    \right\|_2^2
    \right].
    \label{eq:appendix_critic_loss}
\end{equation}
In this way, $G_\mathrm{t}$ supplies the real score, $G_\mathrm{c}$ supplies the fake score, and their score difference provides a distribution-level training signal for $G_\mathrm{s}$. The resulting objective transfers the teacher's generation quality to a few-step student without requiring the student to reproduce the teacher's full denoising trajectory.

\section{Explanation of Decode-Then-Re-Encode Strategy}
\label{app:explanation-dtre}

In Sec.~\ref{sec:framework}, we introduce $\bm{\kappa}^{(k)}$, which carries motion context from the $(k{-}1)$-th chunk via a decode-then-re-encode process: $\bm{\kappa}^{(k)}=\mathcal{E}\left(\mathcal{D}(\hat{\mathbf{x}}^{(k-1)}[-\tau:\,]) \right)$. While directly reusing the tail latents of the $(k{-}1)$-th chunk as the head context of chunk $k$ seems intuitive, it introduces a severe positional-role mismatch due to the design of the 3D causal VAE. Specifically, to maintain image compatibility, the VAE compresses the leading frame of each input sequence purely spatially, while jointly compressing subsequent frames in both space and time. Consequently, the tail latents of chunk $(k{-}1)$ are spatio-temporal aggregates, whereas the head position of chunk $k$ requires a spatially-encoded latent. The proposed decode-then-re-encode operation circumvents this inconsistency by re-encoding the boundary frames under chunk $k$'s positional context, restoring the correct latent role and ensuring seamless temporal continuity.

\section{Optimization Objective for Teacher Model}
\label{app:teacher-loss}

Beyond the standard flow-matching objective $\mathcal{L}_\text{diff}$ defined in Eq.~\ref{eq:diffusion-loss}, we introduce two pixel-level auxiliary losses to further enhance temporal coherence and facial fidelity. 

\textbf{Temporal gradient consistency} penalizes discrepancies in the first-order temporal derivatives between adjacent decoded frames. This promotes natural, smooth motion dynamics without imposing overly restrictive absolute frame-wise constraints: 
\begin{equation}
    \mathcal{L}_\text{temp}^{(k)} = \frac{1}{T-1}\sum_{t=1}^{T-1} \left\| (\hat{\mathbf{V}}^{(k)}_{t+1} - \hat{\mathbf{V}}^{(k)}_t) - (\mathbf{V}^{(k)}_{t+1} - \mathbf{V}^{(k)}_t) \right\|_2^2,
    \label{eq:temp_loss}
\end{equation}
where $\hat{\mathbf{V}}^{(k)} = \mathcal{D}(\hat{\mathbf{x}}^{(k)})$ and $\mathbf{V}^{(k)}$ denote the decoded teacher output and the ground-truth chunk, respectively.

\textbf{Facial region focalization} applies a facial mask $\mathbf{M}_f$ to concentrate gradient on the face region, preventing background regions from dominating optimization:
\begin{equation}
    \mathcal{L}_\text{facial}^{(k)} = \left\| \mathbf{M}_f \odot \bigl(\hat{\mathbf{V}}^{(k)} - \mathbf{V}^{(k)}\bigr) \right\|_2^2.
    \label{eq:facial_loss}
\end{equation}

\section{Dataset Construction Pipeline}
\label{dataset}

Our training corpus is constructed from AVSpeech~\cite{ephrat2018looking}, HDTF~\cite{zhang2021flow}, OpenHumanVid~\cite{li2025openhumanvid}, TalkVid~\cite{chen2025talkvid}, VFHQ~\cite{xie2022vfhq}, and a self-collected corpus. The self-collected subset is gathered from publicly accessible web videos, including creator-shared talking-head clips, interviews, speeches, tutorials, and vlog-style monologue videos retrieved through keyword and category search on open web video platforms. We only retain videos that are publicly available and whose platform terms or creator-provided license permit non-commercial research use, such as Creative Commons-style public sharing licenses or explicit research-friendly terms. Private, login-gated, paywalled, or license-ambiguous videos are excluded. The final corpus contains 217 hours of high-quality synchronized audio-visual pairs. We organize the pipeline into four stages: multi-source collection, temporal standardization, quality filtering, and training tuple construction.

\paragraph{Multi-source Collection.}
Let the raw video pool be
\begin{equation}
    \mathcal{R}
    =
    \mathcal{R}_\text{AVSpeech}
    \cup
    \mathcal{R}_\text{HDTF}
    \cup
    \mathcal{R}_\text{OpenHumanVid}
    \cup
    \mathcal{R}_\text{TalkVid}
    \cup
    \mathcal{R}_\text{VFHQ}
    \cup
    \mathcal{R}_\text{self}.
    \label{eq:appendix_raw_pool}
\end{equation}
Because the same web video may appear in multiple sources, we remove duplicated samples before any model-specific preprocessing. For each raw video $r_i$, we compute a source identifier and a visual hash $h(r_i)$, then keep the first occurrence:
\begin{equation}
    \mathcal{R}_\text{uniq}
    =
    \left\{
    r_i\in\mathcal{R}
    \,\middle|\,
    h(r_i)\notin\{h(r_j)\mid j<i\}
    \right\}.
    \label{eq:appendix_dedup}
\end{equation}
This step avoids over-representing popular identities or repeated clips, which would bias both identity preservation and synchronization evaluation.

\paragraph{Temporal Standardization.}
Each unique video in $\mathcal{R}_\text{uniq}$ is split into semantically coherent clips. For sources with reliable timestamps, we use the provided boundaries; for untrimmed videos, we detect hard scene changes and remove abrupt jump cuts. Each retained clip is normalized to 25 FPS and resized/cropped into a $512\times512$ face-centered sequence. Denoting the standardized clip set by $\mathcal{C}_0$, each clip is represented as
\begin{equation}
    c_i = \left(\mathbf{V}_i,\mathbf{A}_i\right),
    \qquad
    \mathbf{V}_i=\{I_{i,t}\}_{t=1}^{T_i},
    \quad
    \mathbf{A}_i=\{a_{i,u}\}_{u=1}^{U_i},
    \quad
    \mathrm{fps}(\mathbf{V}_i)=25.
    \label{eq:appendix_clip_representation}
\end{equation}
This standardization ensures that an 81-frame training chunk always corresponds to a fixed temporal duration, which is important for aligning Wav2Vec audio features with the latent video tokens used by the chunk-wise generator.

\paragraph{Quality Filtering.}
We apply a cascade of filters to remove samples that are harmful for audio-driven portrait generation. First, a face detector and face parser are used to estimate face visibility and facial masks. Let $d(I_{i,t})\in\{0,1\}$ indicate whether a valid face is detected in frame $t$. The face visibility ratio is
\begin{equation}
    \rho_\text{face}(c_i)
    =
    \frac{1}{T_i}\sum_{t=1}^{T_i} d(I_{i,t}).
    \label{eq:appendix_face_ratio}
\end{equation}
Second, a video quality assessment model~\cite{wu2023exploring} gives a quality score $q_\text{vqa}(c_i)$, and optical/temporal discontinuity checks suppress clips with scene jumps or unstable crops. Third, SyncNet~\cite{chung2016out} measures lip-audio alignment with the same two metrics used in the main text: synchronization confidence $\mathrm{Sync}\text{-}\mathrm{C}(c_i)$ and synchronization distance $\mathrm{Sync}\text{-}\mathrm{D}(c_i)$. The retained set is
\begin{equation}
    \mathcal{C}_\text{clean}
    =
    \left\{
    c_i\in\mathcal{C}_0
    \,\middle|\,
    \rho_\text{face}(c_i)\ge\gamma_f,\;
    q_\text{vqa}(c_i)\ge\gamma_q,\;
    \mathrm{Sync}\text{-}\mathrm{C}(c_i)\ge\gamma_c,\;
    \mathrm{Sync}\text{-}\mathrm{D}(c_i)\le\gamma_d
    \right\},
    \label{eq:appendix_filter_set}
\end{equation}
where $\gamma_f$, $\gamma_q$, $\gamma_c$, and $\gamma_d$ are filtering thresholds. In addition, we use DWPose~\cite{yang2023effective} to estimate whole-body, hand, and face keypoints. Clips are removed when hand keypoints overlap the facial region for a sustained interval, indicating severe hand-over-face occlusion, or when the detected face keypoints are missing or have low confidence across too many sampled frames. We also remove clips with obvious audio corruption. This cascade is deliberately conservative because false positives in the training set directly weaken lip synchronization and may introduce visual artifacts into long-term autoregressive generation.

\paragraph{Training Tuple Construction.}
For each clean clip $c_i\in\mathcal{C}_\text{clean}$, we extract a clean vocal track using an FFmpeg-based audio separation tool and compute frame-aligned audio features with the frozen Wav2Vec encoder followed by the same MLP projection used in the main model. Facial masks are extracted with a pre-trained face parsing model~\cite{yu2021bisenet}:
\begin{equation}
    \mathbf{c}_{a,i}^{(k)}
    =
    \mathrm{MLP}\left(\mathrm{Wav2Vec}\left(\mathbf{A}_i^{(k)}\right)\right),
    \qquad
    \mathbf{M}_{f,i}^{(k)}
    =
    \mathcal{P}\left(\mathbf{V}_i^{(k)}\right),
    \label{eq:appendix_audio_mask}
\end{equation}
where $\mathcal{P}$ denotes the face parsing network. Here, $\mathbf{c}_{a,i}^{(k)}$ is the dataset-indexed version of the audio condition $\mathbf{c}_a^{(k)}$ used in Sec.~\ref{sec:framework}. Training samples are then drawn as consecutive 81-frame chunks. For a clean clip $c_i$, we sample a reference frame $I_{\mathrm{ref},i}$ from the same identity and construct
\begin{equation}
    \mathbb{D}
    =
    \left\{
    \left(
    I_{\mathrm{ref},i},
    \mathbf{V}_i^{(k)},
    \mathbf{c}_{a,i}^{(k)},
    \mathbf{M}_{f,i}^{(k)}
    \right)
    \right\}_{i,k},
    \label{eq:appendix_training_tuple}
\end{equation}
The tuple in Eq.~\ref{eq:appendix_training_tuple} provides the reference identity, target video chunk, audio condition, and facial mask. This keeps dataset annotations separate from method-specific continuity references: $\bm{\kappa}_\text{gt}$ and $\bm{\kappa}_\text{gen}$ are constructed on the fly by the AKD training procedure described in Sec.~\ref{sec:akd}, rather than stored as dataset fields.

\section{Identity Reference Injection Strategies}
\label{app:identity-strategies}

In this section, we will briefly introduce the three identity reference injection strategies we compared in Sec.~\ref{sec:ablation-tre}. 

\textbf{CLIP-based Embedding} follows the semantic branch of~\cite{fei2025skyreels}: the reference image is processed by a frozen CLIP vision encoder $\mathcal{F}_\text{CLIP}(\cdot)$, grid features from the penultimate layer are projected by an MLP into DiT-dimensional image tokens, and these tokens are injected into the DiT through an additional image cross-attention branch whose residual is combined with the audio cross-attention output. Formally, given the reference image $I_\text{ref}$, the identity condition is constructed as
\begin{equation}
    \mathbf{c}_I^\text{CLIP} = \mathrm{MLP}\bigl(\mathcal{F}_\text{CLIP}(I_\text{ref})\bigr) \in \mathbb{R}^{N_\text{CLIP}\times D},
    \label{eq:appendix_clip_embed}
\end{equation}
where $N_\text{CLIP}$ denotes the number of CLIP grid tokens and $D$ is the DiT hidden dimension. For each DiT block, the intermediate hidden state $\mathbf{h}$ is updated as
\begin{equation}
    \mathbf{h}' = \mathbf{h} + \mathrm{CrossAttn}\bigl(\mathbf{h},\,\mathbf{c}_a\bigr) + \mathrm{CrossAttn}\bigl(\mathbf{h},\,\mathbf{c}_I^\text{CLIP}\bigr).
    \label{eq:appendix_clip_inject}
\end{equation}

\textbf{Visual Token Projection} follows the discrete visual-token design in~\cite{deng2025avatarsync}: a frozen Open-MAGVIT2 tokenizer~\cite{luo2024open} quantizes the reference image into spatially ordered discrete visual tokens, and the token embeddings are augmented with positional embeddings before projection to the DiT hidden dimension. Specifically, the tokenizer $\mathcal{Q}(\cdot)$ produces a token sequence
\begin{equation}
    \mathbf{q} = \mathcal{Q}(I_\text{ref}) \in \{1,\dots,V\}^{N_q},
    \label{eq:appendix_vtp_quantize}
\end{equation}
where $V$ is the codebook size and $N_q$ is the number of spatial tokens. The discrete tokens are then embedded, augmented with positional encodings, and projected to the DiT hidden dimension:
\begin{equation}
    \mathbf{c}_I^\text{VTP} = \mathrm{Proj}\bigl(\mathrm{Embed}(\mathbf{q}) + \mathrm{PE}\bigr) \in \mathbb{R}^{N_q\times D},
    \label{eq:appendix_vtp_proj}
\end{equation}
which is fed into the DiT through the same cross-attention interface as in Eq.~\ref{eq:appendix_clip_inject}.

\textbf{Encode-Then-Repeat paradigm} encodes the reference image once with the 3D VAE, repeats the resulting latent along the temporal dimension, and concatenates it with the noisy video latent. Concretely, the source image is first compressed by the frozen 3D-VAE encoder $\mathcal{E}(\cdot)$ into a single latent frame and then replicated $L$ times along the temporal axis to match the chunk length:
\begin{equation}
    \mathbf{c}_I^\text{ETR} = \mathrm{Repeat}_L\bigl(\mathcal{E}(I_\text{ref})\bigr) \in \mathbb{R}^{L\times h\times w\times C},
    \label{eq:appendix_etr_repeat}
\end{equation}
where $L$, $h\times w$, and $C$ follow the latent dimensions defined in Sec.~\ref{sec:framework}. The repeated latent is then fused with the noisy input via channel-wise concatenation, mirroring Eq.~\ref{eq:tre_concat}:
\begin{equation}
    \tilde{\mathbf{x}}_0^{(k)} = \mathrm{Concat}\bigl(\mathbf{x}_0^{(k)},\,\mathbf{c}_I^\text{ETR}\bigr) \in \mathbb{R}^{L\times h\times w\times 2C}.
    \label{eq:appendix_etr_concat}
\end{equation}
In contrast to our TRE (Eq.~\ref{eq:tre}), this paradigm reverses the order of repetition and encoding, which bypasses the VAE's temporal modeling modules and yields a temporally static identity condition.

\section{Additional Ablation Visualizations}
\label{app:vis-ablation}

\begin{figure}[H]
    \centering
    \begin{minipage}{0.5\linewidth}
        \centering
        \includegraphics[width=\linewidth]{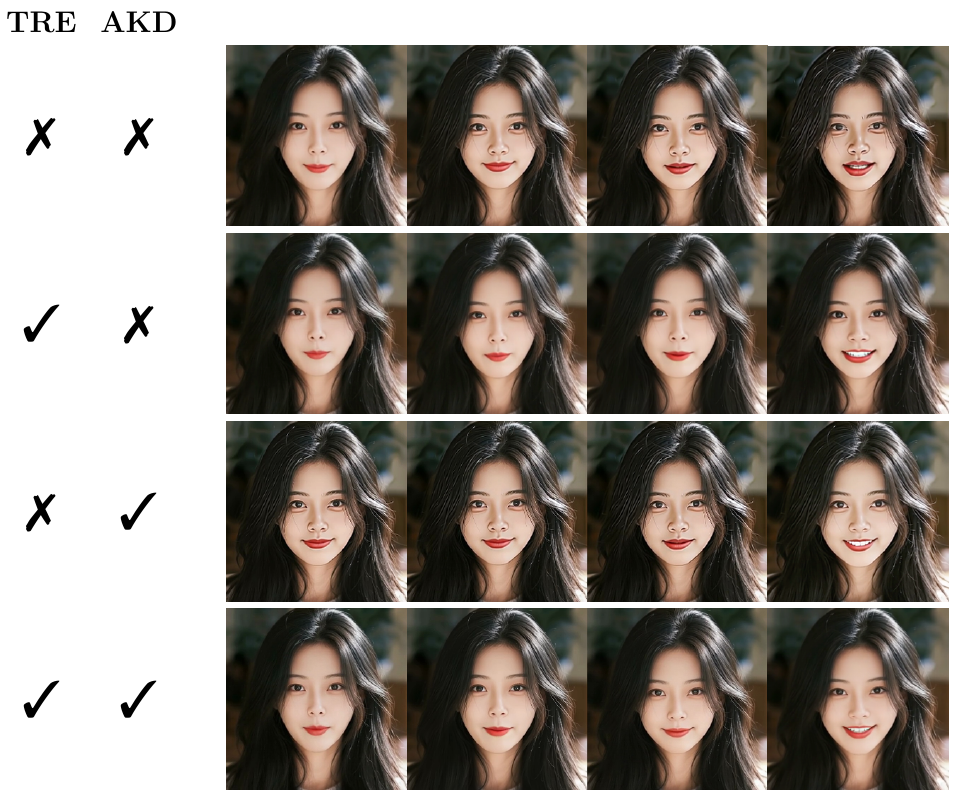}
        \caption{Ablation results of TRE and AKD.}
        \label{fig:ablation_tre_akd}
    \end{minipage}\hspace{0.5em}%
    \begin{minipage}{0.42\linewidth}
        \vspace{0.5em}
        \centering
        \includegraphics[width=\linewidth]{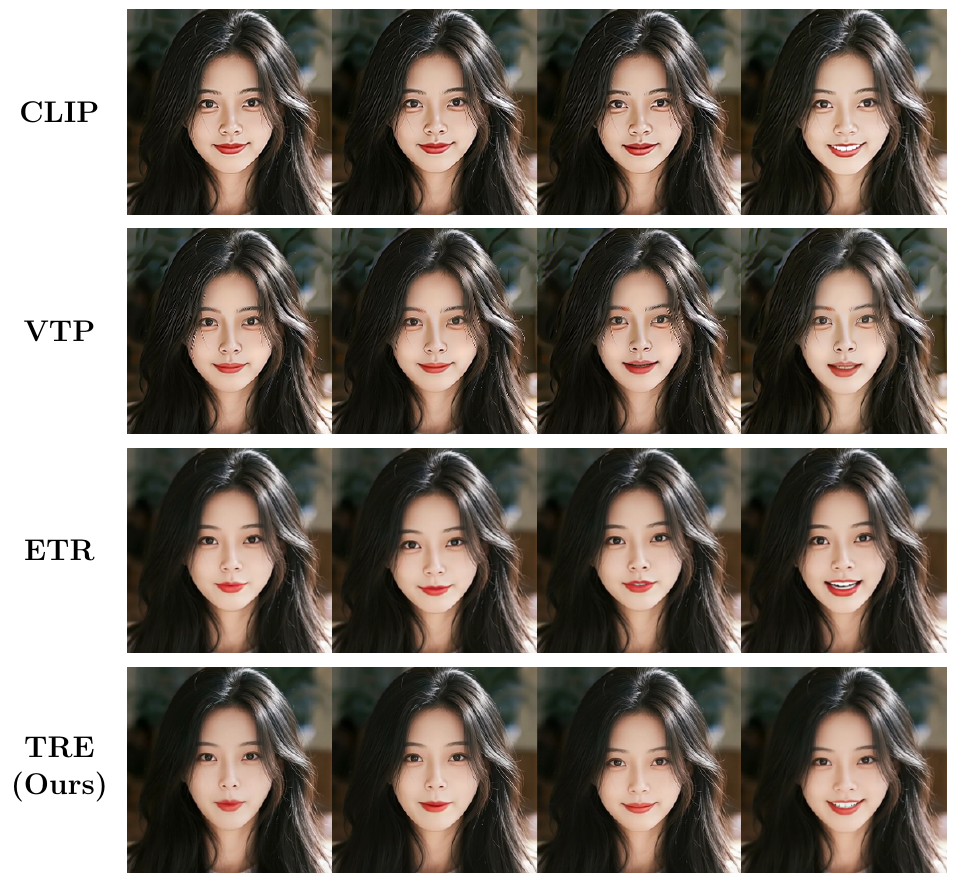}
        \caption{Ablation results of TRE.}
        \label{fig:ablation_tre}
    \end{minipage}
\end{figure}


\begin{figure}[H]
    \centering
    \includegraphics[width=0.7\linewidth]{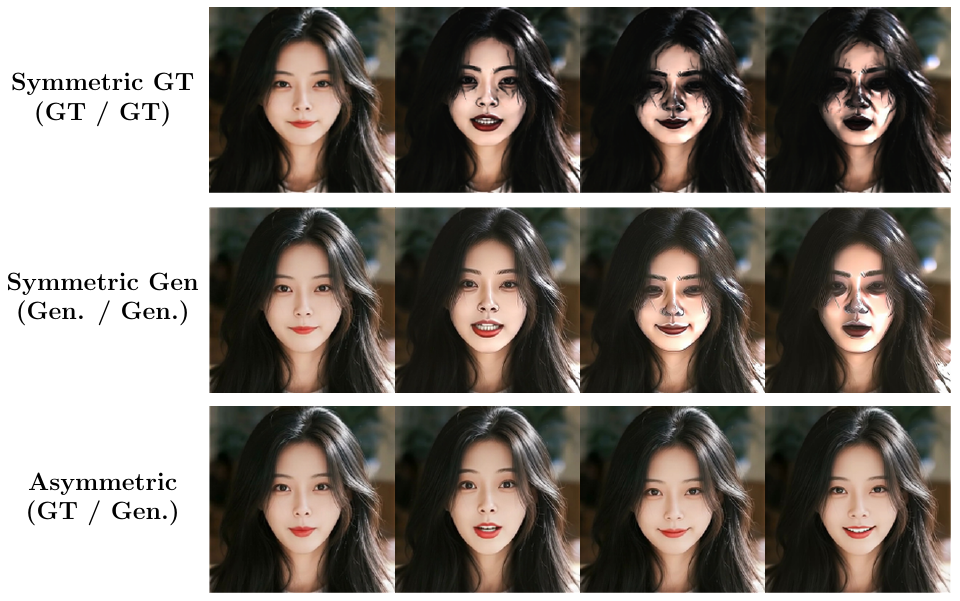}
    \caption{Ablation results of AKD.}
    \label{fig:ablation_akd}
\end{figure}

In this section, we provide qualitative evaluations of our proposed modules. Figure~\ref{fig:ablation_tre_akd} illustrates the visual impact of TRE and AKD; omitting either component leads to noticeable degradation in facial fidelity and expression accuracy. Furthermore, as depicted in Figure~\ref{fig:ablation_tre}, substituting our TRE with alternative identity injection strategies severely compromises temporal stability. This highlights the necessity of strictly aligning the static identity reference with the dynamic audio condition. Finally, Figure~\ref{fig:ablation_akd} visualizes the catastrophic artifact accumulation and structural collapse inherent in symmetric distillation setups (Symmetric GT and Gen.). These dramatic visual failures empirically corroborate the importance of AKD's asymmetric design, which uniquely preserves the teacher's drift-free supervision while maintaining strict train-inference alignment.

\section{Additional Qualitative Results}
\label{app:qualitative}

\begin{figure}[p]
    \vspace*{\fill}
    \centering
    \includegraphics[width=1\linewidth]{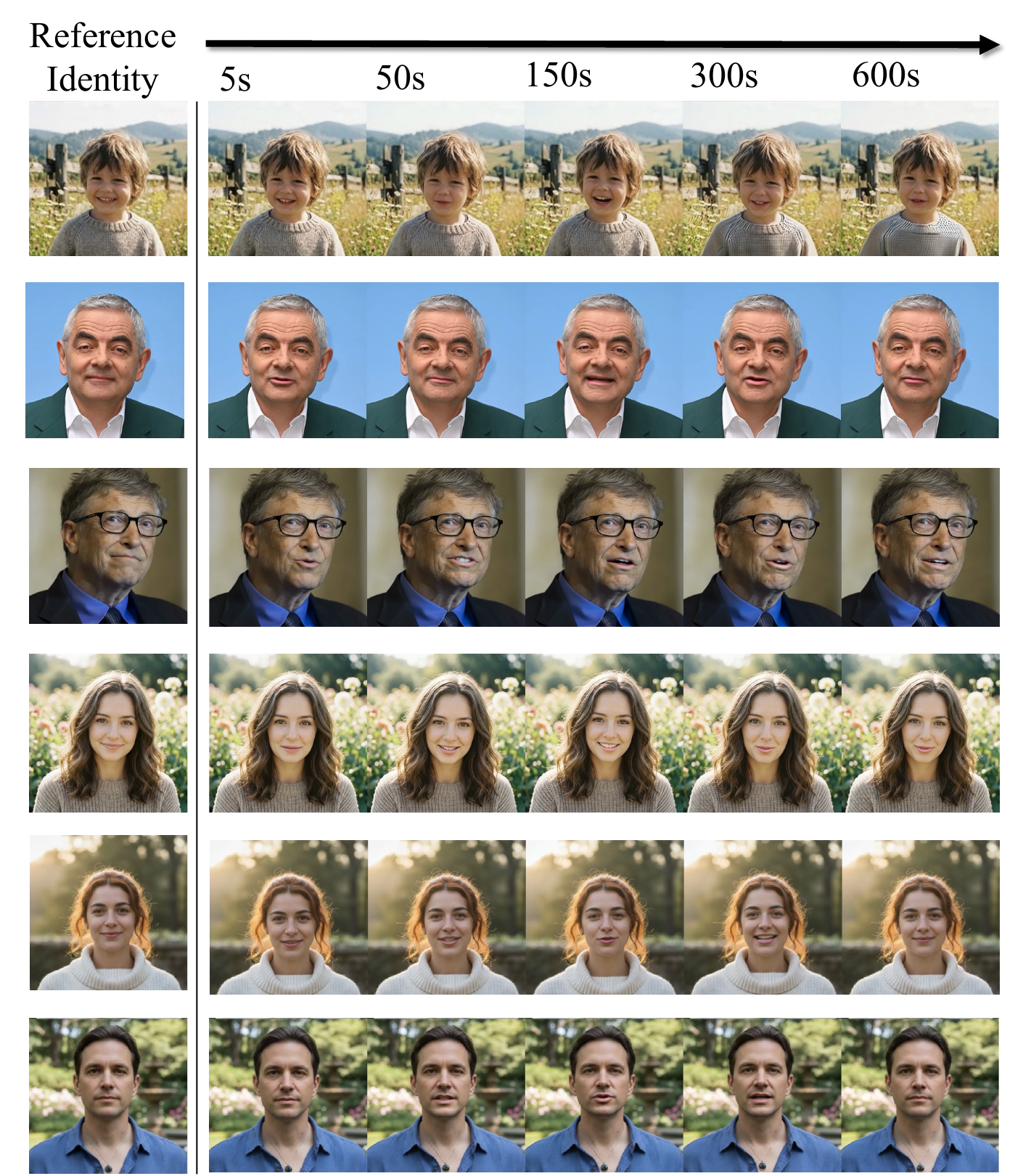}
    \caption{Additional qualitative results of AsymTalker on 600-second long-term video generation. }
    \label{fig:ours_show1}
    \vspace*{\fill}
\end{figure}

\begin{figure}[p]
    \vspace*{\fill}
    \centering
    \includegraphics[width=1\linewidth]{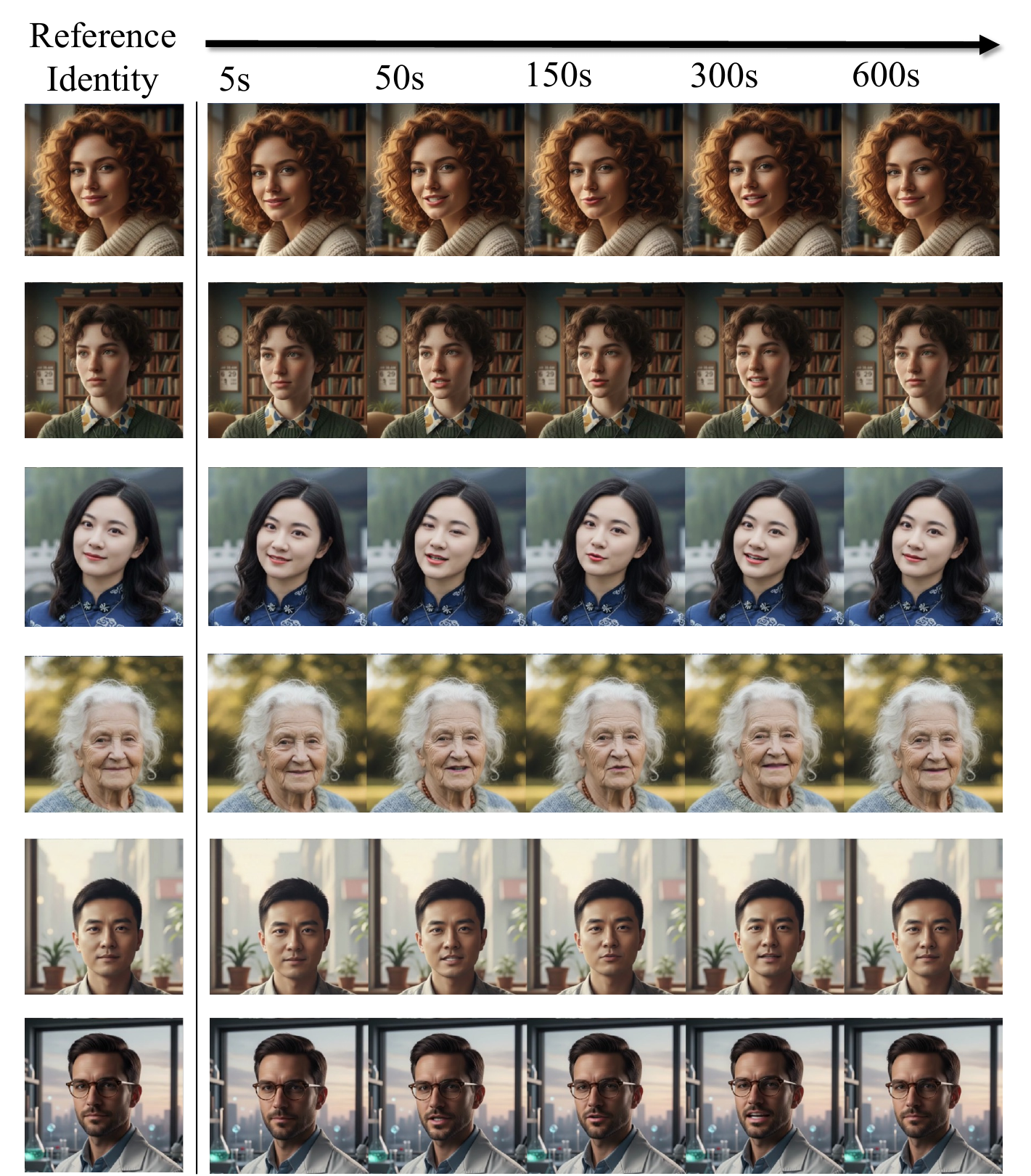}
    \caption{Additional qualitative results of AsymTalker on 600-second long-term video generation.}
    \label{fig:ours_show2}
    \vspace*{\fill}
\end{figure}

\begin{figure}[p]
    \vspace*{\fill}
    \centering
    \includegraphics[width=1\linewidth]{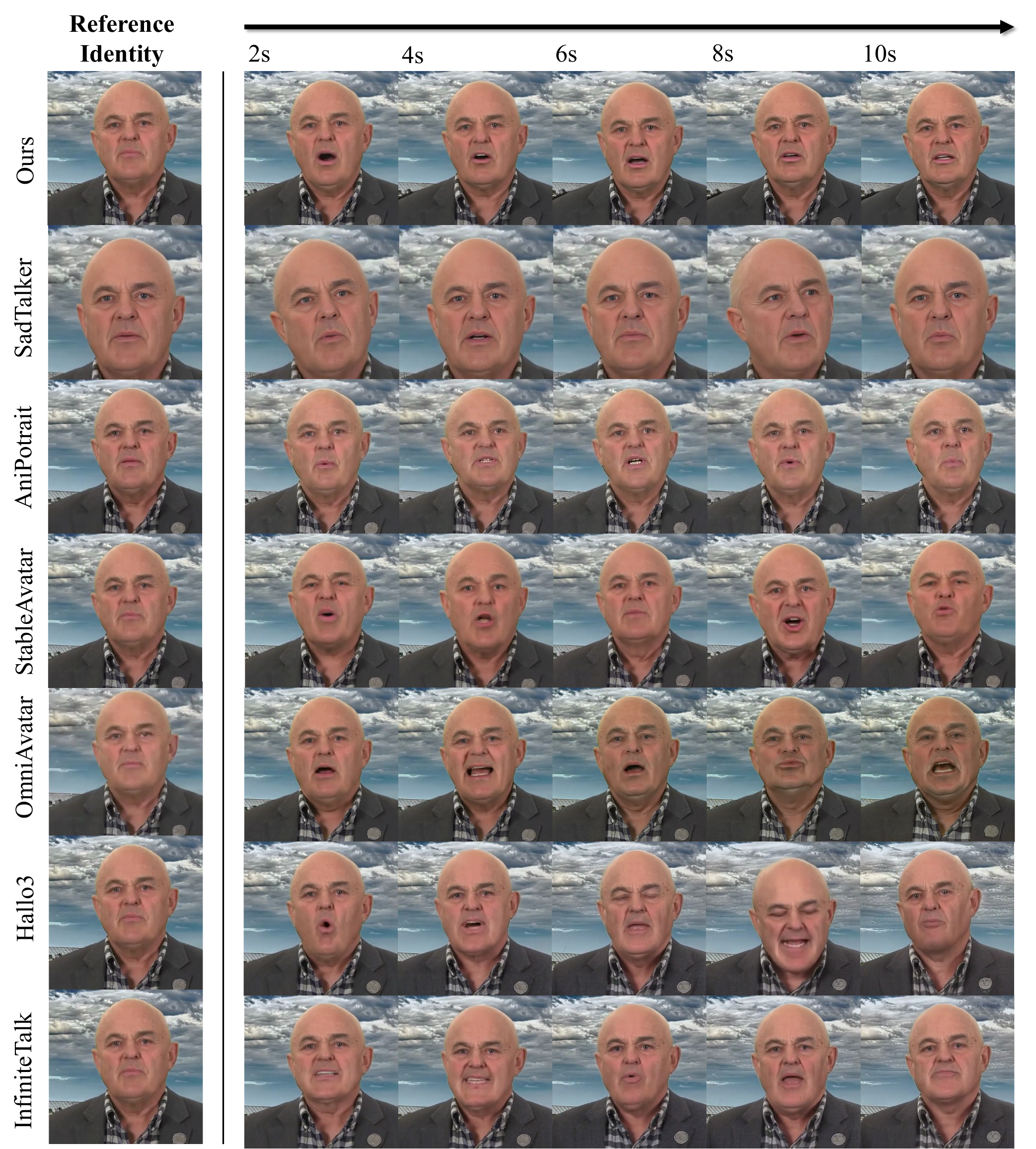}
    \caption{Qualitative comparison on the HDTF dataset. }
    \label{fig:compare1}
    \vspace*{\fill}
\end{figure}

\begin{figure}[p]
    \vspace*{\fill}
    \centering
    \includegraphics[width=1\linewidth]{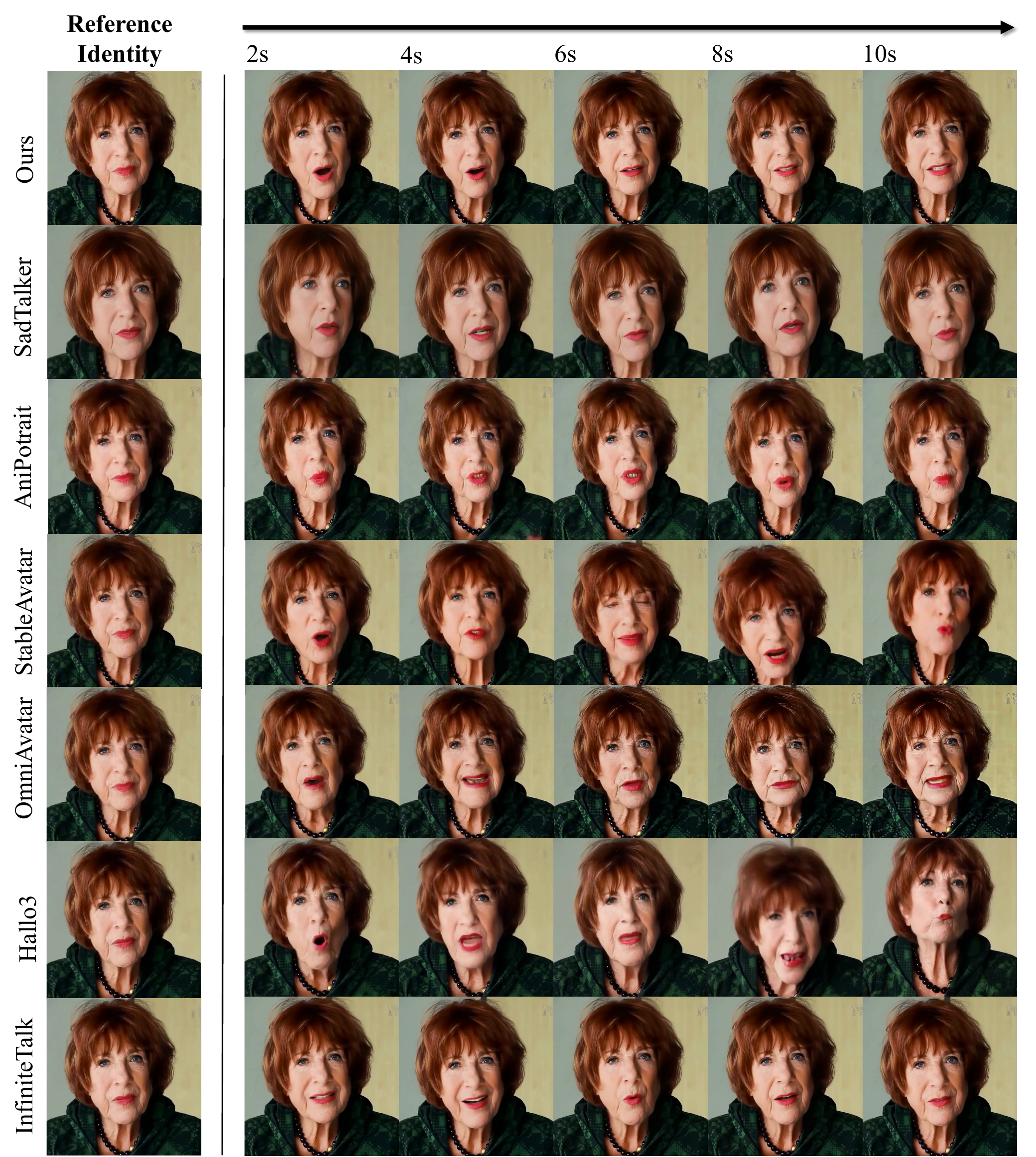}
    \caption{Qualitative comparison on the VFHQ dataset.}
    \label{fig:compare2}
    \vspace*{\fill}
\end{figure}

In this section, we present further qualitative evaluations. These include extended long-term video generation results produced by AsymTalker (Figures~\ref{fig:ours_show1}-\ref{fig:ours_show2}), as well as comprehensive visual comparisons against state-of-the-art baselines (Figures~\ref{fig:compare1}-\ref{fig:compare2}).

\section{Broader Impacts}
\label{app:broader-impacts}

AsymTalker targets long-term, real-time talking head generation from a reference image and driving audio. Potential positive impacts include more accessible video communication, lower-cost educational and creative production, assistive avatar interfaces for users who cannot easily appear on camera, and research tools for studying audio-visual generation and synchronization. The method also improves efficiency, which may reduce the marginal compute cost of generating long-form talking-head videos once a model has been trained.

At the same time, talking-head synthesis is a dual-use technology. Higher visual fidelity and longer temporal stability can increase risks of impersonation, non-consensual identity use, misinformation, harassment, and erosion of trust in online media. These risks are especially salient when generated videos are presented without disclosure or when a person's identity is used without consent. We therefore frame the system as a research contribution rather than a deployment-ready product, and we pair the release with the safeguards described below.

\section{Safeguards for Responsible Release}
\label{app:safeguards}

We adopt a staged release strategy. The anonymous repository provides code and reproduction instructions for research review, while large-scale data assets are released only after additional license, privacy, and safety auditing. The released materials will be accompanied by research-use terms that prohibit impersonation, non-consensual identity synthesis, deceptive political or commercial use, harassment, and other harmful applications.

For data, we exclude private, login-gated, paywalled, and license-ambiguous sources; remove clips with obvious corruption or unsafe content during filtering; and retain only publicly accessible videos whose platform terms or creator-provided licenses permit research use. For generated media, users are instructed to obtain consent from identity owners, avoid using the system to imitate real people without authorization, and clearly disclose synthetic outputs. When releasing model checkpoints or processed data, we will include documentation of intended use, restricted use, and takedown/contact procedures for content owners.

\section{Asset Usage and Documentation}
\label{app:asset-usage}

Our work uses existing datasets, pretrained models, and tools including AVSpeech~\cite{ephrat2018looking}, HDTF~\cite{zhang2021flow}, OpenHumanVid~\cite{li2025openhumanvid}, TalkVid~\cite{chen2025talkvid}, VFHQ~\cite{xie2022vfhq}, Wan2.1~\cite{wan2025wan}, Wav2Vec~\cite{baevski2020wav2vec}, SyncNet~\cite{chung2016out}, DWPose~\cite{yang2023effective}, and face parsing tools~\cite{yu2021bisenet}. We cite the original creators and use these assets according to licenses, public terms, or research-use conditions that permit our non-commercial research setting. Sources whose terms are unclear or incompatible with research use are excluded from the training corpus.

For new assets introduced by this work, the main paper documents the model architecture, training objectives, inference procedure, and evaluation protocol; Appendix~\ref{dataset} documents dataset construction and filtering; and the anonymous GitHub repository provides code-level details and reproduction instructions. The dataset is large, so its public release is separated from the initial code release and will follow the audit procedure described in Appendix~\ref{app:safeguards}.


\end{document}